\documentclass[10pt,twocolumn,letterpaper]{article}

\usepackage[pagenumbers]{cvpr} 

\usepackage{tikz}
\usetikzlibrary{arrows.meta,calc,positioning}
\usepackage{pifont}
\usepackage{array}

\usepackage{amsthm}
\newtheorem{proposition}{Proposition}

\usepackage{colortbl}
\definecolor{oursbg}{RGB}{252,234,238}
\newcommand{\na}{--}

\newcommand{\imgslot}[2]{%
  \IfFileExists{#1}%
    {\includegraphics[width=\linewidth,height=0.7\linewidth,keepaspectratio]{#1}}%
    {\fbox{\parbox[c][\dimexpr0.7\linewidth-2\fboxsep-2\fboxrule\relax][c]%
      {\dimexpr\linewidth-2\fboxsep-2\fboxrule\relax}%
      {\centering\scriptsize\color{gray}#2\\[2pt]\texttt{\detokenize{#1}}}}}}

\definecolor{cvprblue}{rgb}{0.21,0.49,0.74}
\usepackage[pagebackref,breaklinks,colorlinks,allcolors=cvprblue]{hyperref}

\AtBeginDocument{%
}

\def\confName{CVPR}
\def\confYear{2027}

\title{Pretraining Linear-Time Foundation Models on Information-Thin SAR Images}

\author{Ke Wang\textsuperscript{1,}\thanks{Equal contribution.}\quad
Jiale Liu\textsuperscript{2,}\footnotemark[1]\quad
Lei Chen\textsuperscript{1}\quad
Nanzhe Wang\textsuperscript{3}\quad
Xiaofeng Ai\textsuperscript{1}\quad
Feng Zhao\textsuperscript{1}\quad
Xiaoyi Pan\textsuperscript{1,}\thanks{Corresponding author.}\\[0.6em]
{\small\textsuperscript{1}National University of Defense Technology\quad
\textsuperscript{2}The University of Edinburgh\quad
\textsuperscript{3}Heriot-Watt University}\\[0.3em]
{\tt\small\href{mailto:wk_wkk0310@163.com}{wk\_wkk0310@163.com}\quad
\href{mailto:Jiale.Liu@ed.ac.uk}{Jiale.Liu@ed.ac.uk}\quad
\href{mailto:mrpanxy@nudt.edu.cn}{mrpanxy@nudt.edu.cn}}}

\begin{document}
\maketitle

\begin{abstract}
Most synthetic aperture radar (SAR) foundation models follow the optical recipe: Transformers, often initialized from ImageNet, pretrained with masked image modeling, plus designs that suppress speckle as interference. This recipe pays the quadratic cost of softmax attention on large scenes and does not ask where the task information lies. We argue that SAR scenes with homogeneous clutter and sparse targets are \emph{information-thin}: clutter reduces to a few local statistics that a recurrence can update, speckle creates no information shared across regions, and task evidence concentrates on sparse scattering structures. A fixed-size recurrent state can thus summarize the clutter and hold the few structures, so a linear-time encoder can suffice if pretraining teaches it what to keep. We therefore pretrain SARVim, a bidirectional \mbox{Mamba-3} encoder, from scratch on MillionSAR, 1.07M images from 29 public SAR datasets. Physics-aware masked pretraining masks tokens by scatter saliency and reconstructs structure and clutter statistics instead of speckled pixels, while a scatter-aware spatial gate in every block emphasizes scattering structures. SARVim-B outperforms comparable-size foundation models on seven of eight detection, classification, and segmentation benchmarks, and its per-token cost stays nearly constant from 640 to 2048 pixels, against up to 6.8$\times$ growth for global-attention Transformers. Code, models, and MillionSAR will be released.%
\end{abstract}

\section{Introduction}
\label{sec:intro}

\begin{figure}[t]
  \centering
  \includegraphics[width=\columnwidth]{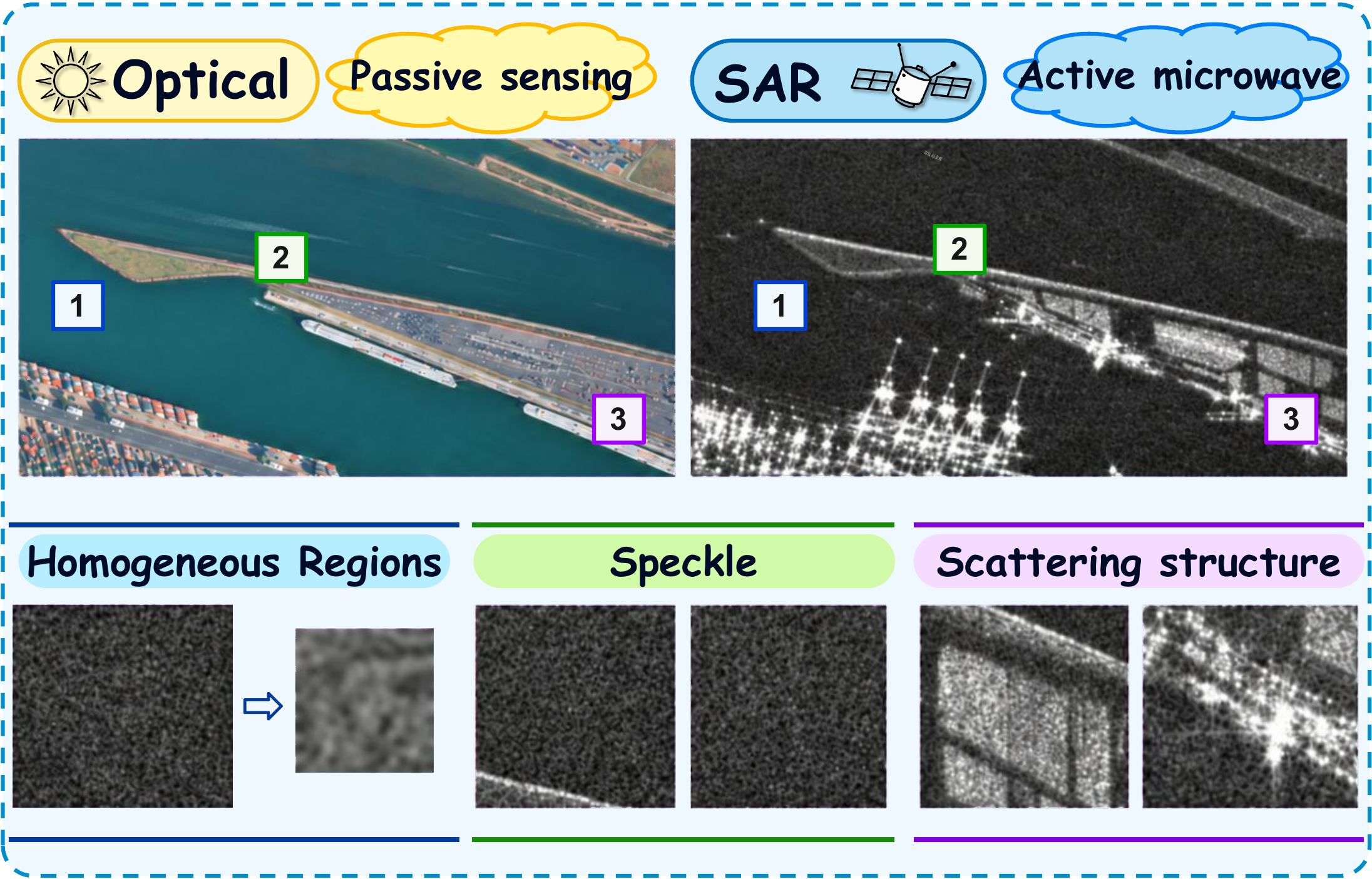}
  \caption{\textbf{SAR images are information-thin.} (1) Clutter is summarized by local statistics, (2) speckle carries no shared information, and (3) task evidence concentrates on scattering structures.}
  \label{fig:resides}
\end{figure}

Synthetic aperture radar (SAR) images the ground day and night, through cloud, making it a core sensor for maritime surveillance, disaster assessment, and reconnaissance~\cite{moreira2013tutorial}. Unlike an optical pixel, a SAR pixel records a coherent sum of backscattered waves (\autoref{fig:resides}). A man-made target appears as a few dominant scattering centers~\cite{araujo2022scattering}, with layover and shadow from the side-looking geometry, whereas homogeneous clutter is covered by speckle, whose standard deviation equals the local mean in single-look intensity~\cite{goodman1976speckle}. Most pixel-level variation of a SAR image is thus speckle, independent of the scene. Task evidence concentrates on sparse scattering structures and a few clutter statistics; we call such images \emph{information-thin}.

Many state-of-the-art SAR foundation models~\cite{li2025saratrx,li2026saratrxv2,liu2025sarmae} nevertheless follow the optical recipe of initializing a Transformer from ImageNet and pretraining it on SAR images with masked image modeling. This recipe fits information-thin images poorly in two ways. First, it treats speckle as interference, to be countered with speckle-robust gradient targets~\cite{li2024sarjepa,li2025saratrx}, auxiliary denoising~\cite{du2025summit}, or speckle injection~\cite{liu2025sarmae}, and does not ask where the task information lies. Second, softmax attention\footnote{Here, attention means global softmax self-attention, which keeps every token and can retrieve any token by content.} compares every pair of tokens, at quadratic cost, although most tokens are speckled background. The recipe is self-reinforcing: Transformers train poorly on small datasets~\cite{dosovitskiy2021vit}, ImageNet initialization compensated while SAR data were scarce~\cite{li2025saratrx}, and public masked-image-modeling checkpoints are mostly Transformers~\cite{he2022mae,zhang2023hivit,tian2023itpn}. Prior work has thus dropped one part at a time: SAR-JEPA~\cite{li2024sarjepa} and SUMMIT~\cite{du2025summit} pretrain ViTs without ImageNet but retain softmax attention, whereas SAMBA~\cite{wang2026samba} adopts Mamba but starts from ImageNet. The scarcity has eased: pooling 29 public datasets yields MillionSAR, 1.07M images close in size to ImageNet-1K, half of them from datasets released since 2025. Both the encoder and its pretraining can now be designed around the information-thinness of SAR images.

Designing both requires knowing what a SAR foundation model must keep as it reads an image. We argue that the answer lies in information-thinness, which rests on three properties (\autoref{sec:theory}): (i) homogeneous clutter is described by a few local statistics that a recurrence can update, (ii) speckle creates no information shared across regions, and (iii) task evidence concentrates on sparse scattering structures. The data agree. In MillionSAR, the coefficient of variation in homogeneous blocks is 0.32 against 0.028 in COCO, and strong scattering occupies a median of 16.5\% of the tokens. The annotated targets of four detection benchmarks cover a median of 1.7\% of an image against 38.6\% in COCO. On SSDD, the detection errors of SARVim-B likewise follow the number and size of these sparse targets rather than the scene type, except in crowded harbors. Two consequences follow. First, a linear-time encoder can suffice: the fixed-size state of a recurrence costs only the task information it discards, however large the pixel variation, and a summary of the clutter and the few structures discards little. Second, such a summary must be learned, so the same properties tell pretraining where to mask, what to reconstruct, and which tokens to emphasize.

Accordingly, we pretrain SARVim from scratch on MillionSAR with a linear-time bidirectional \mbox{Mamba-3}~\cite{lahoti2026mamba3} encoder (\autoref{sec:method}). SARVim emphasizes scattering structures in masking, targets, and gating: it masks scatter-salient tokens more often, so the model must infer them from their visible surroundings. It reconstructs gain-invariant gradients, speckle texture, and saliency instead of speckled pixels, and a scatter-aware spatial gate amplifies scattering tokens in every block. Without ImageNet initialization or softmax attention, SARVim-B leads Base-scale foundation models on seven of eight detection, classification, and segmentation benchmarks. Linear-time models are commonly expected to trade accuracy for efficiency~\cite{jelassi2024repeat,arora2024zoology}. On SAR, we find this trade-off avoidable: SARVim-B outperforms a ViT-B with the same pretraining on all three tasks, while its per-token cost stays nearly constant from 640 to 2048 pixels.
Our contributions are as follows:
\begin{itemize}
  \item We characterize SAR scenes with homogeneous clutter and sparse targets as information-thin through an information-theoretic analysis, support it with the pixel statistics of 1.07M SAR images, the target area of four detection benchmarks, and the detection errors on SSDD, and derive principles for pretraining linear-time encoders.
  \item We introduce MillionSAR, a pretraining corpus of 1.07M SAR images pooled from 29 public datasets in five frequency bands, with resolutions from 0.1 to 60~m.
  \item We propose SARVim, a linear-time \mbox{Mamba-3} foundation model pretrained from scratch on MillionSAR with saliency-guided masking, physics-aware targets, and a scatter-aware spatial gate.
\end{itemize}

\section{Related Work}
\label{sec:related}

\paragraph{SAR Pretraining and Initialization.} SAR pretraining has grown from target chips to million-scale corpora. SAR-JEPA~\cite{li2024sarjepa} predicts multi-scale gradients of masked regions, pretraining a ViT on 95K images without ImageNet. SARATR-X~\cite{li2025saratrx} and SARATR-X-v2~\cite{li2026saratrxv2} pretrain ImageNet-initialized HiViT and iTPN with multi-scale gradient and multi-receptive-field structural targets, respectively. SUMMIT~\cite{du2025summit} pretrains a ViT-B from scratch on 570K images with auxiliary denoising, edge, and scattering-point tasks. SARMAE~\cite{liu2025sarmae} injects speckle into masked autoencoding and builds SAR-1M. SAMBA~\cite{wang2026samba} pretrains an ImageNet-initialized bidirectional Mamba with scatter-guided masking on 186K images. Multimodal Earth-observation models treat SAR as one input among several~\cite{fuller2023croma,wang2025copernicusfm,yang2026mars,li2026babelrs}. Task-specific SAR models already exploit scattering centers in target recognition~\cite{araujo2022scattering}, scattering keypoints in ship detection~\cite{fu2021skgnet}, and scattering characteristics in aircraft classification~\cite{sun2022saracd}. In natural images, training from scratch matches ImageNet initialization given longer schedules or in-domain self-supervision~\cite{he2019rethinking,zoph2020rethinking,hernandez2021scaling,elnouby2021large}, whereas SARATR-X still finds it helpful with 186K SAR images; we forgo it with MillionSAR. Most SAR pretraining designs above make models robust to speckle, and the scattering-point task of SUMMIT and the masking of SAMBA each target scatterers with a single component; we instead design our masking, target, gate, and encoder around the information-thinness of SAR images, whose task information lies on sparse scattering structures.

\paragraph{Selective State Space Models.} Structured state space models (SSMs)~\cite{gu2022s4} process a sequence in linear time by recursively updating a fixed-size state, which matches the recursive summary of homogeneous clutter (\autoref{prop:clutter}). Their selective variant Mamba~\cite{gu2024mamba} makes the update input-dependent, so the state can restart at region changes and keep the few scattering structures while summarizing the background; Mamba-3~\cite{lahoti2026mamba3} adds a complex-valued update equivalent to a data-dependent rotary embedding, which writes relative position into the state and thus helps it keep where these structures lie. The linear cost also suits large SAR scenes, for which softmax attention grows quadratically with the number of tokens. Vision backbones such as Vim~\cite{zhu2024vim}, VMamba~\cite{liu2024vmamba}, and MambaVision~\cite{hatamizadeh2025mambavision} reach competitive ImageNet accuracy, and MAP~\cite{liu2025map}, ARM~\cite{ren2025arm}, and RoMA~\cite{wang2025roma} pretrain Mamba autoregressively; we instead keep masked autoencoding and, following information-thinness, focus the Mamba encoder on scattering structures: a scatter-aware spatial gate amplifies their tokens in every block, and during pretraining the visible tokens are sorted by keep score, with the salient ones last in the forward scan.

\section{Information-Thin SAR Images}
\label{sec:theory}

This section asks where the task information of a SAR image lies. We define information-thinness, relate it to clutter, speckle, and scattering structures with evidence from data, and derive principles for pretraining.

\begin{figure*}[t]
  \centering
  \includegraphics[width=\textwidth]{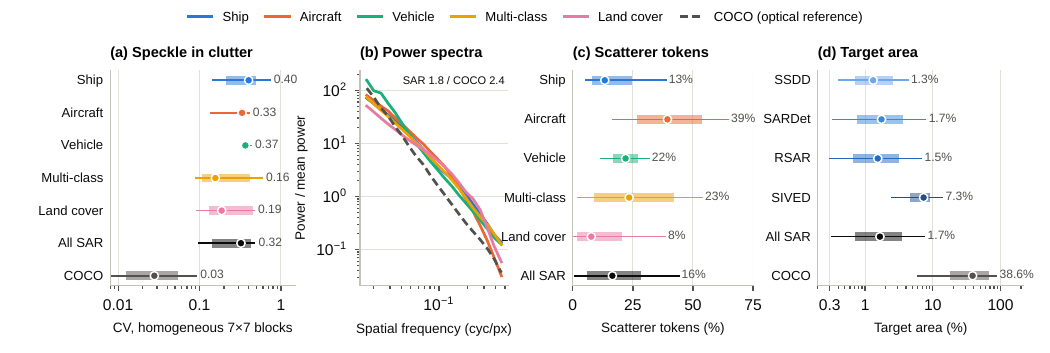}
  \caption{\textbf{Pixel statistics of MillionSAR and target area of SAR benchmarks.} COCO val2017 is the optical reference. \textbf{(a)}~Coefficient of variation in the most homogeneous $7{\times}7$ blocks. \textbf{(b)}~Median power spectrum, normalized by the mean power. \textbf{(c)}~Fraction of tokens whose scatter saliency is at least 0.5 (SAR only). \textbf{(d)}~Fraction of the image covered by the union of target boxes, with All SAR pooling the four detection benchmarks. Bars: interquartile range; whiskers: 10th--90th percentiles; dots: median.}
  \label{fig:pixel}
\end{figure*}

\subsection{Formalization}
\label{sec:theory_formal}

\paragraph{Problem Setting.} A SAR image $X$ is the $L$-look intensity of the coherent sum of scatterer responses and clutter. We use the multiplicative model only for fully developed speckle in homogeneous clutter:
\begin{equation}
  X(u)=R(u)\,N(u),\qquad N(u)\sim\mathrm{Gamma}(L,1/L),
  \label{eq:speckle}
\end{equation}
where $\mathbb{E}[N]=1$ and $\mathrm{Var}(N)=1/L$~\cite{goodman1976speckle,oliver1998understanding}, while pixels dominated by deterministic scatterers follow other statistics.

\paragraph{Information-Thinness.} A backbone reads the image as a token sequence in scan order, and a cut at position $t$ splits it into a prefix $X_A$ and a suffix $X_B$. A scene is \emph{information-thin} for a task $Y$ when a compact summary $S=f(X_A)$ loses little task information (\autoref{prop:summary}; proofs in \autoref{sec:supp_proofs}). We hypothesize that SAR scenes with homogeneous background and sparse targets are information-thin; this does not mean low pixel entropy or an easy task, and the hypothesis can fail in harbors with bright infrastructure.

\begin{proposition}[Summary loss]
  \label{prop:summary}
  For a discrete label $Y$, any summary $S=f(X_A)$, and Bayes-optimal probabilistic predictors evaluated by log-loss $\mathcal{L}^*$,
  \begin{equation}
    \mathcal{L}^*(S,X_B)-\mathcal{L}^*(X_A,X_B)=I(Y;X_A\mid S,X_B).
    \label{eq:summary}
  \end{equation}
\end{proposition}

A forward recurrence reads the prefix only through its state $S_t$ at the cut, so its outputs $Z_B$ after the cut satisfy
\begin{equation}
  I(X_A;Z_B\mid X_B)\le I(X_A;S_t\mid X_B)\le H(S_t)\le B_{\mathrm{state}},
  \label{eq:state}
\end{equation}
where the last two steps assume finite precision and, for bidirectional or multi-layer models, $B_{\mathrm{state}}$ counts every message that crosses the cut. Softmax attention has no such bottleneck, so by \autoref{prop:summary}, a recurrent model can match it only if some summary within $B_{\mathrm{state}}$ makes \autoref{eq:summary} small; whether such a summary can be learned is a separate question. The relevant quantity is thus task information rather than pixel correlation.

\subsection{Clutter and Speckle}
\label{sec:theory_bg}

\begin{proposition}[Background summary]
  \label{prop:speckle}\label{prop:clutter}\label{sec:theory_background}
  (i)~In a homogeneous region with known $L$ and constant mean $\mu$, if $X_1,\dots,X_n$ are independent with $X_i\sim\mathrm{Gamma}(L,\mu/L)$, then $T_n=(n,\sum_{i=1}^{n}X_i)$ is sufficient for $\mu$ and can be updated online. (ii)~If $A$ and $B$ partition the pixels and the speckle fields $N_A$ and $N_B$ are independent of each other and of $R$, then $I(X_A;X_B)\le I(R_A;R_B)$.
\end{proposition}

The clutter mean thus has a recursive summary, and speckle creates no shared information of its own~\cite{cover2006elements}, so the background part of a summary can be small and computed by a recurrence. Neither result implies that the shared information is small or that recognition needs no long-range relations. The summary must restart at region changes and exclude targets, which input-dependent updates allow.

\paragraph{Evidence.} Compared with 4,952 grayscale COCO images~\cite{lin2014coco}, the statistics of MillionSAR (\autoref{sec:data}) show that much of the pixel variation of SAR images is speckle (\autoref{fig:pixel}a,b): the coefficient of variation in homogeneous blocks is 0.32 against 0.028, and the power spectrum decays more slowly, with a median slope of 1.8 against 2.4.

\subsection{Scattering Structures}
\label{sec:theory_st}

Property (iii) concerns the rest of the image: a useful state must also hold the scattering structures, shadows, and spatial relations still unresolved at the cut; since \autoref{eq:state} bounds only the total state, SARVim need not keep them apart from the background summary. If a task must keep $K_{\mathrm{live}}$ structures, each located at one of $n_{\mathrm{tok}}$ tokens and described by $b_a$ bits, an explicit event list costs
\begin{equation}
  B_{\mathrm{events}}\lesssim K_{\mathrm{live}}\bigl(\lceil\log_2 n_{\mathrm{tok}}\rceil+b_a\bigr)+B_{\mathrm{rel}},
  \label{eq:events}
\end{equation}
where $B_{\mathrm{rel}}$ encodes their relations. This cost, an upper bound on the optimal state, grows with the number of structures but only logarithmically with that of tokens, so a fixed-size state suffices when structures are few. Selective SSMs, with input-dependent updates, suit such scenes in linear time; convolutions and local attention exploit this sparsity only within their receptive fields.

\paragraph{Evidence.} Scattering structures are sparse in MillionSAR: the scatter saliency of \autoref{eq:saliency} reaches 0.5 for a median of 16.5\% of an image's tokens, from 7.7\% for land-cover scenes to 39.4\% for aircraft (\autoref{fig:pixel}c). Annotated targets are sparser still: their boxes cover a median of 1.7\% of an image over four detection benchmarks, from 1.3\% on SSDD to 7.3\% on SIVED, against 38.6\% on COCO val2017, which annotates more categories (\autoref{fig:pixel}d).

\subsection{Implications for Pretraining}
\label{sec:theory_impl}

The three properties place the task information on scattering structures and a few clutter statistics, which tells a pretext task where to look. By (iii), masking should fall on the scattering structures, so that the model must infer them from their surroundings. By (i) and (ii), the reconstruction target should describe structures and local clutter statistics rather than speckled pixels. By (ii) and (iii), each block should emphasize tokens on scattering structures over speckle-dominated ones.

\section{Pretraining SARVim}
\label{sec:method}

\begin{figure*}[t]
  \centering
  \includegraphics[width=\linewidth]{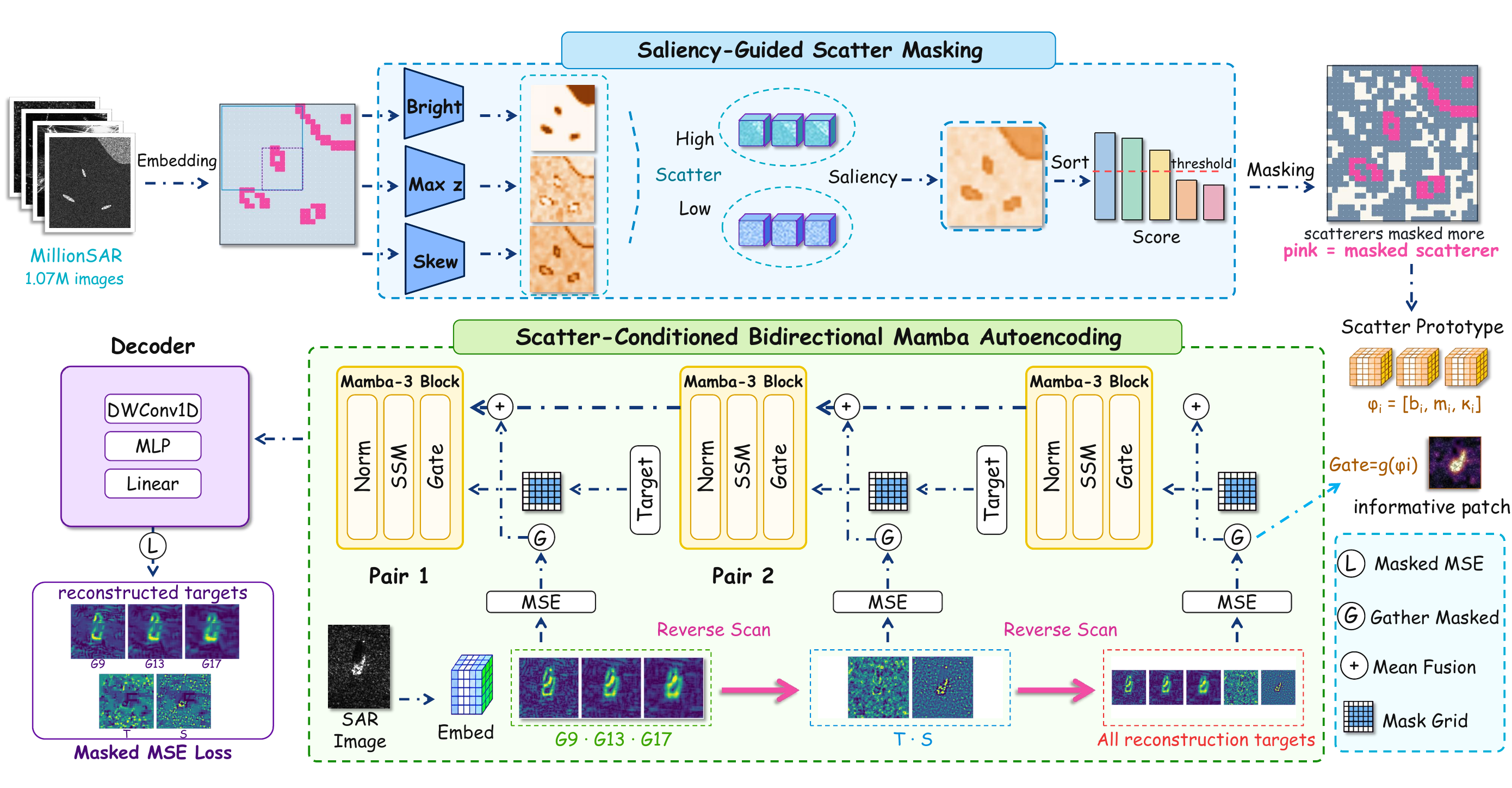}
  \caption{\textbf{Overview of SARVim.} It has two parts: (a) saliency-guided masking, which combines per-token statistics $\hat{b}_i$, $\hat{m}_i$, and $\hat{\kappa}_i$ into the scatter saliency $s_i$ and masks salient tokens more often; and (b) a bidirectional Mamba-3 autoencoder, whose encoder reads the visible tokens, sorted by keep score, with the scatter-aware spatial gate, and whose lightweight decoder reconstructs physics-aware targets ($G_9$, $G_{13}$, $G_{17}$, $T$, $S$) for the masked tokens. The pretrained encoder transfers to detection, classification, and segmentation.}
  \label{fig:method}
\end{figure*}

This section turns these principles into SARVim, pretrained from scratch. We pool public datasets into MillionSAR, build a linear-time encoder that emphasizes scattering structures, and pretrain it by masked autoencoding with saliency-guided masking and physics-aware targets (\autoref{fig:method}).

\subsection{MillionSAR}
\label{sec:data}

MillionSAR, the corpus that lets SARVim be pretrained without ImageNet, pools 29 public SAR datasets; supplementary \autoref{fig:corpora} compares it with the corpora of previous SAR foundation models. Nine of its sources, with 202K images, appear in none of them. It contains 1,066,055 images with 228 gigapixels. By content, 43.7\% of the images show land-cover scenes, 23.7\% multi-class scenes, 22.5\% vehicles, 7.9\% ships, and 2.2\% aircraft. The images are spaceborne (77.5\%), airborne (20.0\%), or simulated (2.5\%), and cover the S, C, X, Ku, and Ka bands at resolutions from about 0.1 to 60~m. We remove the images in the official test splits of all downstream benchmarks.

\subsection{SARVim Encoder}
\label{sec:method_backbone}

Since information-thin images need only a summary of the background and a few scattering structures (\autoref{sec:theory_st}), we build the encoder on a recurrence, whose cost is linear in the number of tokens.

\paragraph{Tokenization.} SARVim embeds a single-channel input $X\in\mathbb{R}^{224\times224}$ with a $16{\times}16$ convolution of stride 8 into a $27{\times}27$ grid of $n_{\mathrm{tok}}=729$ tokens. Following Vim~\cite{zhu2024vim}, we insert a learnable CLS token in the middle of the token sequence and add learnable absolute position embeddings.

\paragraph{Mamba-3 Layers.} Each layer is a pre-norm residual block with a Mamba-3 mixer~\cite{lahoti2026mamba3}, which computes per token a value $x_t$, gate $z_t$, projections $B_t$, $C_t$, step size $\Delta_t>0$, decay rate $A_t<0$, trapezoidal coefficient $\lambda_t\in(0,1)$, and rotation rates $\theta_t$, and updates the state of each head by
\begin{equation}
  h_t=\alpha_t h_{t-1}+(1-\lambda_t)\Delta_t\alpha_t\,\bar{B}_{t-1}x_{t-1}^{\top}+\lambda_t\Delta_t\,\bar{B}_t x_t^{\top},
  \label{eq:mamba3}
\end{equation}
where $\alpha_t=\exp(\Delta_t A_t)$ is the data-dependent decay and $\bar{B}_t=R(\Phi_t)B_t$ rotates half the state dimensions pairwise by accumulated angles $\Phi_t=\sum_{s\le t}\Delta_s\theta_s$, equivalent to a complex-valued state transition. The layer reads the state with equally rotated $\bar{C}_t$ and gates the output by $\mathrm{SiLU}(z_t)$.

\paragraph{Bidirectional Scanning.} Since a forward recurrence reads the prefix through its state (\autoref{eq:state}), we arrange 24 layers into 12 pairs, where one layer scans the token sequence forward and the other backward, and average their outputs and residual streams. Every token depends on the whole image after the first pair.

\paragraph{Scatter Saliency.} For each token $i$, we measure how strongly its patch departs from homogeneous clutter. Let $\hat{b}_i$, $\hat{m}_i$, and $\hat{\kappa}_i$ denote the fraction of pixels above the 95th percentile of the image, the maximum $z$-score, and the skewness of the pixel values in the patch, each mapped linearly to $[0,1]$. We define the scatter saliency as:
\begin{equation}
  s_i=\mathrm{clip}\bigl(0.4\,\hat{b}_i+0.35\,\hat{m}_i+0.25\,\hat{\kappa}_i,\,0,\,1\bigr).
  \label{eq:saliency}
\end{equation}
Under \autoref{eq:speckle}, homogeneous clutter has a fixed skewness of $2/\sqrt{L}$, whereas a strong scatterer produces a few very bright pixels that raise all three statistics.

\paragraph{Scatter-Aware Spatial Gate.} By (ii) and (iii), speckle-dominated tokens carry little task information, so we make the distinction between scattering structures and background explicit in every block and gate the mixer output $o_i$ of each token before the residual addition:
\begin{equation}
  o_i\leftarrow o_i\odot(1+\alpha s_i)\odot\bigl(1+0.3\,g(\phi_i)\bigr),
  \label{eq:modulation}
\end{equation}
where the scatter-aware gain $1+\alpha s_i$ amplifies tokens on scattering structures, and the physics gate $g$ maps the token statistics $\phi_i$ to channel-wise weights. Since neither factor falls below 1, no token is suppressed.

\subsection{Physics-Aware Masked Pretraining}
\label{sec:method_pretrain}

\paragraph{Saliency-Guided Masking.} By (iii), we mask scatterers more often, using a continuous score rather than binary labels and block sampling~\cite{wang2026samba}. With $\tilde{s}_i$ the saliency normalized to $[0,1]$ per image and $c_i\in[0,1]$ the contrast between token $i$ and its eight neighbors, the masking score is:
\begin{equation}
  p_i=r_{\mathrm{bg}}+\tilde{s}_i\,(r_{\mathrm{sc}}-r_{\mathrm{bg}})+0.2\,(c_i-0.5),
  \label{eq:maskprob}
\end{equation}
where $r_{\mathrm{sc}}>r_{\mathrm{bg}}$ favor scatterers. After a small random dither, which gives $\hat{p}_i$, we mask the 75\% of tokens with the highest $\hat{p}_i$ (\autoref{sec:supp_impl}) and feed the visible tokens by decreasing keep score $1-\hat{p}_i$, so that the forward layer of each pair reads background tokens first and the most salient visible tokens last.

\paragraph{Physics-Aware Reconstruction Targets.} By (ii), speckle adds no information shared across regions, so reconstructing raw pixels spends capacity on noise; we reconstruct statistics instead. For structure boundaries, we compute multi-scale ratio gradients~\cite{li2024sarjepa,li2025saratrx}: with $m^{\mathrm{L}}_r$, $m^{\mathrm{R}}_r$, $m^{\mathrm{T}}_r$, and $m^{\mathrm{B}}_r$ the means over the left, right, top, and bottom halves of a $(2r{+}1)\times(2r{+}1)$ window around pixel $u$,
\begin{equation}
  G_r(u)=\Bigl[\log^2\frac{m^{\mathrm{L}}_r(u)}{m^{\mathrm{R}}_r(u)}+\log^2\frac{m^{\mathrm{T}}_r(u)}{m^{\mathrm{B}}_r(u)}\Bigr]^{1/2},
  \label{eq:gf}
\end{equation}
with $r\in\{9,13,17\}$; the ratios make $G_r$ invariant to multiplicative gain, which varies across MillionSAR sources. Beyond gradient features, which previous SAR masked models reconstruct alone~\cite{li2024sarjepa,li2025saratrx,wang2026samba}, two further channels cover the background and the scatterers. The speckle texture $T(u)=\log\bigl(1+\sigma_7(u)/\mu_7(u)\bigr)$, the log-scaled coefficient of variation in $7{\times}7$ windows, describes the clutter statistics of (i); under \autoref{eq:speckle} it is fixed by $L$ in fully developed speckle and rises with heterogeneity. The saliency $S(u)$, a per-image normalized difference of Gaussians, highlights bright scatterers and, unlike $s_i$, dark shadows, the structures of (iii). A lightweight decoder regresses the five channels of masked tokens with an MSE loss.

\section{Experiments}
\label{sec:exp}

This section compares SARVim with previous methods, tests the predictions of the analysis, ablates its design, and measures its efficiency.

\subsection{Experimental Setup}
\label{sec:exp_setup}

\paragraph{Benchmarks and Baselines.} We evaluate on eight public benchmarks covering detection, classification, and segmentation. For detection, we use SSDD~\cite{zhang2021ssdd}, SARDet-100K~\cite{li2024sardet100k}, RSAR~\cite{zhang2025rsar}, and SIVED~\cite{lin2023sived}. For classification, we use MSTAR~\cite{ross1998mstar} and SAR-ACD~\cite{sun2022saracd}. For segmentation, we use AIR-PolSAR-Seg-2.0~\cite{wang2025airpolsarseg2} and WHU-OPT-SAR~\cite{li2022mcanet} with SAR only. We compare with SAR foundation models and pretraining methods, with Earth-observation foundation models evaluated on SAR, and with task-specific models on each benchmark, using the results reported in their papers.

\paragraph{Implementation Details.} We pretrain SARVim-B and SARVim-L on MillionSAR for 600 epochs on 32 GH200 GPUs with AdamW~\cite{loshchilov2019adamw}, a batch size of 4,096, and bf16 mixed precision, using random crops, flips, and rotations without color augmentation; the supplementary material lists the hyperparameters of pretraining and detection.

\subsection{Main Results}
\label{sec:exp_sota}

\begin{table*}[t]
  \centering
  \caption{\textbf{Performance comparison (\%) on horizontal and oriented object detection.} The best and second-best results of each foundation-model group are in \textbf{bold} and \underline{underlined}; \na: not reported.}
  \label{tab:sota}
  \footnotesize
  \setlength{\tabcolsep}{3.5pt}
  \begin{minipage}[t]{0.49\textwidth}
    \centering
    \begin{tabular}{@{}llccc@{}}
      \multicolumn{5}{c}{(a) SSDD~\cite{zhang2021ssdd}} \\
      \toprule
      Method & Backbone & mAP & AP$_{50}$ & AP$_{75}$ \\
      \midrule
      CS$^n$Net~\cite{chen2023csnnet}    & CSPDarkNet-53 & 64.9 & 97.1 & \na \\
      MSFA~\cite{li2024sardet100k}       & VAN-B         & \na  & 97.9 & \na \\
      \midrule
      SARATR-X~\cite{li2025saratrx}      & HiViT-B       & 67.5 & 97.3 & 83.5 \\
      SUMMIT~\cite{du2025summit}         & ViT-B         & 70.4 & 96.7 & \underline{86.3} \\
      SARMAE~\cite{liu2025sarmae}        & ViT-B         & 68.1 & \na  & \na \\
      SAMBA~\cite{wang2026samba}         & Mamba         & \underline{71.3} & \underline{97.8} & 85.6 \\
      SARATR-X-v2~\cite{li2026saratrxv2} & iTPN-B        & 71.0 & 96.5 & 85.6 \\
      \rowcolor{oursbg}
      SARVim (ours)                      & Mamba-3-B     & \textbf{74.2} & \textbf{98.5} & \textbf{91.1} \\
      \midrule
      ViTP~\cite{li2025vitp}             & ViT-L         & 70.8 & \underline{97.8} & 86.6 \\
      SARMAE~\cite{liu2025sarmae}        & ViT-L         & 69.3 & \na  & \na \\
      SARATR-X-v2~\cite{li2026saratrxv2} & iTPN-L        & \underline{71.8} & 97.0 & \underline{87.6} \\
      \rowcolor{oursbg}
      SARVim (ours)                      & Mamba-3-L     & \textbf{74.0} & \textbf{98.7} & \textbf{89.8} \\
      \bottomrule
    \end{tabular}
  \end{minipage}\hfill
  \begin{minipage}[t]{0.49\textwidth}
    \centering
    \begin{tabular}{@{}llccc@{}}
      \multicolumn{5}{c}{(b) SARDet-100K~\cite{li2024sardet100k}} \\
      \toprule
      Method & Backbone & mAP & AP$_{50}$ & AP$_{75}$ \\
      \midrule
      MSFA~\cite{li2024sardet100k}       & ConvNeXt-B    & 56.4 & 88.2 & 61.5 \\
      \midrule
      MaRS~\cite{yang2026mars}           & SwinV2-B      & 55.4 & \na  & 61.4 \\
      SARATR-X~\cite{li2025saratrx}      & HiViT-B       & 57.3 & 88.7 & 62.8 \\
      SUMMIT~\cite{du2025summit}         & ViT-B         & 57.0 & 89.9 & 62.7 \\
      SARMAE~\cite{liu2025sarmae}        & ViT-B         & 57.9 & \na  & \na \\
      SAMBA~\cite{wang2026samba}         & Mamba         & 59.7 & 89.1 & \underline{68.0} \\
      SARATR-X-v2~\cite{li2026saratrxv2} & iTPN-B        & \underline{63.4} & \textbf{92.5} & \textbf{70.1} \\
      \rowcolor{oursbg}
      SARVim (ours)                      & Mamba-3-B     & \textbf{64.2} & \underline{90.0} & 66.5 \\
      \midrule
      ViTP~\cite{li2025vitp}             & ViT-L         & 59.7 & \na  & \na \\
      BabelRS~\cite{li2026babelrs}       & ViT-L         & 63.3 & \underline{91.7} & \na \\
      SARMAE~\cite{liu2025sarmae}        & ViT-L         & 63.1 & \na  & \na \\
      SARATR-X-v2~\cite{li2026saratrxv2} & iTPN-L        & \textbf{68.7} & \textbf{93.9} & \textbf{75.3} \\
      \rowcolor{oursbg}
      SARVim (ours)                      & Mamba-3-L     & \underline{64.5} & 90.1 & \underline{68.5} \\
      \bottomrule
    \end{tabular}
  \end{minipage}

  \vspace{6pt}
  \begin{minipage}[t]{0.49\textwidth}
    \centering
    \begin{tabular}{@{}llccc@{}}
      \multicolumn{5}{c}{(c) RSAR~\cite{zhang2025rsar}} \\
      \toprule
      Method & Backbone & mAP & AP$_{50}$ & AP$_{75}$ \\
      \midrule
      RoI Transformer~\cite{ding2019roitrans}      & ResNet-50   & 35.0 & 67.0 & 32.7 \\
      OrientedFormer~\cite{zhao2024orientedformer} & ResNet-50   & 37.5 & 69.8 & 35.7 \\
      GSDet~\cite{ding2025gsdet}                   & ResNet-50   & 35.6 & 68.0 & 33.5 \\
      O2-RTDETR~\cite{ding2026o2rtdetr}            & ResNet-18vd & 39.5 & 73.1 & 36.9 \\
      \midrule
      SARMAE~\cite{liu2025sarmae}                  & ViT-B       & \na  & 66.8 & \na \\
      SARATR-X-v2~\cite{li2026saratrxv2}           & iTPN-B      & \underline{38.7} & \underline{74.2} & \underline{35.7} \\
      \rowcolor{oursbg}
      SARVim (ours)                                & Mamba-3-B   & \textbf{44.7} & \textbf{82.6} & \textbf{42.8} \\
      \midrule
      ViTP~\cite{li2025vitp}                       & ViT-L       & \na  & 72.3 & \na \\
      SARMAE~\cite{liu2025sarmae}                  & ViT-L       & \na  & 72.2 & \na \\
      SARATR-X-v2~\cite{li2026saratrxv2}           & iTPN-L      & \underline{38.7} & \underline{77.3} & \underline{34.2} \\
      \rowcolor{oursbg}
      SARVim (ours)                                & Mamba-3-L   & \textbf{44.7} & \textbf{82.3} & \textbf{42.9} \\
      \bottomrule
    \end{tabular}
  \end{minipage}\hfill
  \begin{minipage}[t]{0.49\textwidth}
    \centering
    \begin{tabular}{@{}llccc@{}}
      \multicolumn{5}{c}{(d) SIVED~\cite{lin2023sived}} \\
      \toprule
      Method & Backbone & mAP & AP$_{50}$ & AP$_{75}$ \\
      \midrule
      Rotated Faster R-CNN~\cite{yang2020rotatedfrcnn}  & ResNet-50 & 53.1 & 97.8 & 51.0 \\
      Gliding Vertex~\cite{xu2021glidingvertex}         & ResNet-50 & 56.1 & 98.3 & 59.6 \\
      Oriented RepPoints~\cite{li2022orientedreppoints} & ResNet-50 & 60.2 & 99.1 & 70.7 \\
      SVDDD~\cite{wang2025svddd}                        & ResNet-50 & \na  & \na  & 74.9 \\
      \midrule
      SAMBA~\cite{wang2026samba}                        & Mamba     & \underline{67.6} & \underline{97.3} & \underline{82.5} \\
      \rowcolor{oursbg}
      SARVim (ours)                                     & Mamba-3-B & \textbf{68.7} & \textbf{97.8} & \textbf{82.6} \\
      \midrule
      \rowcolor{oursbg}
      SARVim (ours)                                     & Mamba-3-L & 67.9 & 98.0 & 81.4 \\
      \bottomrule
    \end{tabular}
  \end{minipage}
\end{table*}

\paragraph{Horizontal and Oriented Object Detection.} Among Base-scale foundation models, SARVim-B is best on all four detection benchmarks and exceeds the best task-specific detectors (\autoref{tab:sota}). On SSDD, its 74.2~mAP surpasses even the best Large-scale encoder, SARATR-X-v2 with iTPN-L, by 2.4 points; the margin is 1.5 in AP$_{50}$ but 3.5 in AP$_{75}$, so the gain lies mainly in precise localization. In oriented detection, SARVim-B leads SARATR-X-v2 on RSAR by 6.0~mAP, 8.4~AP$_{50}$, and 7.1~AP$_{75}$, the largest margin among the four benchmarks, and surpasses SAMBA on SIVED by 1.1~mAP without the ImageNet initialization that SAMBA uses. At the Large scale, SARVim-L, with 168.0M parameters against 288--304M for the other Large-scale encoders, is best on SSDD and RSAR and second on SARDet-100K. Since detection relies on the few tokens covering scattering structures, these results demonstrate the effectiveness of SARVim, which emphasizes these structures in its masking, reconstruction targets, and spatial gate.

\begin{table*}[t]
  \centering
  \caption{\textbf{Performance comparison (\%) on classification and segmentation.} Marks as in \autoref{tab:sota}.}
  \label{tab:clsseg}
  \footnotesize
  \setlength{\tabcolsep}{3.5pt}
  \begin{minipage}[t]{0.49\textwidth}
    \centering
    \begin{tabular}{@{}llc@{}}
      \multicolumn{3}{c}{(a) MSTAR~\cite{ross1998mstar}, 5-shot} \\
      \toprule
      Method & Backbone & Accuracy \\
      \midrule
      DKTS-N~\cite{zhang2022dktsn}              & \na       & 72.3 \\
      ConvT~\cite{wang2022convt}                & \na       & 75.2 \\
      CRID~\cite{wang2023crid}                  & Swin      & 74.6 \\
      BIDFC~\cite{zhai2022bidfc}                & ResNet-18 & 90.3 \\
      PD~\cite{zhang2024pd}                     & A-ConvNet & 70.2 \\
      \midrule
      SAR-JEPA~\cite{li2024sarjepa}             & ViT-B     & 56.5 \\
      SARATR-X~\cite{li2025saratrx}             & HiViT-B   & 95.9 \\
      SAMBA~\cite{wang2026samba}                & Mamba     & 95.0 \\
      SARATR-X-v2~\cite{li2026saratrxv2}        & iTPN-B    & \underline{96.3} \\
      \rowcolor{oursbg}
      SARVim (ours)                             & Mamba-3-B & \textbf{96.9} \\
      \midrule
      SARATR-X-v2~\cite{li2026saratrxv2}        & iTPN-L    & \textbf{98.7} \\
      \rowcolor{oursbg}
      SARVim (ours)                             & Mamba-3-L & \underline{95.8} \\
      \bottomrule
    \end{tabular}
  \end{minipage}\hfill
  \begin{minipage}[t]{0.49\textwidth}
    \centering
    \begin{tabular}{@{}llc@{}}
      \multicolumn{3}{c}{(b) SAR-ACD~\cite{sun2022saracd}, 30\%} \\
      \toprule
      Method & Backbone & Accuracy \\
      \midrule
      ResNet-50~\cite{he2016resnet}             & ResNet-50 & 59.7 \\
      Swin Transformer~\cite{liu2021swin}       & Swin      & 63.2 \\
      ViT~\cite{dosovitskiy2021vit}             & ViT-B     & 80.0 \\
      \midrule
      CROMA~\cite{fuller2023croma}              & ViT-B     & 89.0 \\
      Copernicus-FM~\cite{wang2025copernicusfm} & ViT-B     & \underline{92.6} \\
      SUMMIT~\cite{du2025summit}                & ViT-B     & 84.3 \\
      SARMAE~\cite{liu2025sarmae}               & ViT-B     & \textbf{95.1} \\
      SAMBA~\cite{wang2026samba}                & Mamba     & 81.1 \\
      \rowcolor{oursbg}
      SARVim (ours)                             & Mamba-3-B & 83.2 \\
      \midrule
      SARMAE~\cite{liu2025sarmae}               & ViT-L     & \textbf{95.6} \\
      \rowcolor{oursbg}
      SARVim (ours)                             & Mamba-3-L & \underline{86.0} \\
      \bottomrule
    \end{tabular}
  \end{minipage}

  \vspace{6pt}
  \begin{minipage}[t]{0.49\textwidth}
    \centering
    \begin{tabular}{@{}llccc@{}}
      \multicolumn{5}{c}{(c) AIR-PolSAR-Seg-2.0~\cite{wang2025airpolsarseg2}} \\
      \toprule
      Method & Backbone & PA & mPA & mIoU \\
      \midrule
      PSPNet~\cite{zhao2017pspnet}               & ResNet-101       & 92.4 & 87.6 & 82.2 \\
      DeepLabV3+~\cite{chen2018deeplabv3plus}    & Xception         & 93.1 & 88.8 & 83.1 \\
      PointRend~\cite{kirillov2020pointrend}     & ResNet-101       & 93.2 & 89.5 & 84.5 \\
      DANet~\cite{fu2019danet}                   & ResNet-101       & 93.9 & 91.1 & 86.0 \\
      TerraSegNet~\cite{wijaya2025terrasegnet}   & EfficientNetV2-S & 97.3 & \na  & 92.1 \\
      \midrule
      SARATR-X-v2~\cite{li2026saratrxv2}         & iTPN-B           & \underline{96.0} & \underline{94.0} & \underline{90.8} \\
      \rowcolor{oursbg}
      SARVim (ours)                              & Mamba-3-B        & \textbf{97.6} & \textbf{96.0} & \textbf{93.4} \\
      \midrule
      SARATR-X-v2~\cite{li2026saratrxv2}         & iTPN-L           & \textbf{98.0} & \textbf{97.2} & \textbf{95.3} \\
      \rowcolor{oursbg}
      SARVim (ours)                              & Mamba-3-L        & \underline{97.9} & \underline{96.8} & \underline{94.3} \\
      \bottomrule
    \end{tabular}
  \end{minipage}\hfill
  \begin{minipage}[t]{0.49\textwidth}
    \centering
    \begin{tabular}{@{}llcc@{}}
      \multicolumn{4}{c}{(d) WHU-OPT-SAR~\cite{li2022mcanet}, SAR only} \\
      \toprule
      Method & Backbone & PA & mIoU \\
      \midrule
      OCRNet~\cite{yuan2020ocrnet}               & HRNetV2-W48      & 78.9 & 44.8 \\
      SETR~\cite{zheng2021setr}                  & ViT-L            & 79.1 & 44.5 \\
      SegFormer~\cite{xie2021segformer}          & MiT-B4           & 79.3 & 45.3 \\
      Mask2Former~\cite{cheng2022mask2former}    & ResNet-101       & 79.3 & 46.8 \\
      PIDNet~\cite{xu2023pidnet}                 & PIDNet-S         & 78.4 & 44.3 \\
      TransUNet~\cite{chen2024transunet}         & ResNet50+ViT     & 80.1 & 47.1 \\
      \midrule
      SARATR-X-v2~\cite{li2026saratrxv2}         & iTPN-B           & \underline{79.8} & \underline{46.1} \\
      \rowcolor{oursbg}
      SARVim (ours)                              & Mamba-3-B        & \textbf{81.6} & \textbf{47.7} \\
      \midrule
      SARATR-X-v2~\cite{li2026saratrxv2}         & iTPN-L           & \underline{79.8} & \underline{47.1} \\
      \rowcolor{oursbg}
      SARVim (ours)                              & Mamba-3-L        & \textbf{81.4} & \textbf{47.8} \\
      \bottomrule
    \end{tabular}
  \end{minipage}
\end{table*}
\paragraph{Target Classification.} We train a linear classifier on the frozen encoder and report top-1 accuracy averaged over 10 runs on MSTAR with five images per class and on SAR-ACD with 30\% of the training set (\autoref{tab:clsseg}). On MSTAR, SARVim-B reaches 96.9\%, the best among Base-scale foundation models and 0.6 points above SARATR-X-v2 with iTPN-B under the same linear probe, and exceeds SARATR-X and SAMBA, which fine-tune part or all of the encoder. On SAR-ACD, the only benchmark where it is not best, SARVim-B reaches 83.2\%, above SAMBA but below SARMAE, which also fine-tunes the encoder. At the Large scale, SARVim-L trails SARATR-X-v2 and SARMAE but raises the accuracy on SAR-ACD to 86.0\%. Frozen SARVim features thus separate vehicle classes, whose scattering structures its pretraining emphasizes, whereas fine-grained aircraft types appear harder to separate without fine-tuning.

\paragraph{Semantic Segmentation.} Among Base-scale foundation models, SARVim-B is best on both segmentation benchmarks and exceeds the best task-specific models (\autoref{tab:clsseg}), leading SARATR-X-v2 with iTPN-B by 2.6~mIoU on AIR-PolSAR-Seg-2.0 and 1.6~mIoU on WHU-OPT-SAR. Its mPA gain of 2.0 points on AIR-PolSAR-Seg-2.0 indicates that the improvement extends across classes, and on WHU-OPT-SAR it also exceeds SARATR-X-v2 with iTPN-L. At the Large scale, SARVim-L is best on WHU-OPT-SAR and trails SARATR-X-v2 on AIR-PolSAR-Seg-2.0. SARVim also suits dense prediction: land-cover classes differ in the clutter statistics of (i), whose summary restarts at boundaries. Qualitative examples are in supplementary Figures~\ref{fig:supp_qual_seg} and~\ref{fig:supp_qual_det}.

\paragraph{Testing Information-Thinness.}\label{sec:exp_theory} We test two predictions of the analysis on the 232 SSDD test images~\cite{zhang2021ssdd}: detection errors should grow with the number of targets in an image (\autoref{eq:events}) and rise where harbor clutter and many targets coincide, outside the conditions of \autoref{prop:clutter}. We group the detections of the fine-tuned SARVim-B by the official inshore and offshore lists, the number of ships, and the box side (supplementary \autoref{tab:supp_ssdd_groups}). AP falls, with non-overlapping intervals, from 79.4 for one ship to 69.5 for five or more, consistent with \autoref{eq:events}, and from 78.2 for boxes above 40 pixels to 68.8 below 20 pixels. The 6.4~AP gap between inshore and offshore detection reflects density rather than the scene: inshore images contain 3.7 ships on average against 2.0 offshore, and at matched density the two scenes differ by at most 0.4~AP for up to four ships. Only for five or more ships, where harbor clutter and many targets coincide, does the gap widen to 7.8~AP, consistent with the second prediction. These results support the analysis: the detection error follows the number of scattering structures, and clutter matters only in crowded harbors. The supplementary material adds a pixel-level study.

\subsection{Ablation Study}
\label{sec:exp_ablation}

The principles of \autoref{sec:theory_impl} prescribe where to mask, what to reconstruct, and which tokens to emphasize. For masking, we rely on the ablation of SAMBA, where scatter-guided masking improves a 27M ImageNet-initialized Mamba over uniform random masking by 4.7 to 7.7 points on MSTAR~\cite{ross1998mstar}, FUSAR-Ship~\cite{hou2020fusar}, and SAR-ACD~\cite{sun2022saracd}; we combine these datasets for the linear probe (Cls) of \autoref{tab:ablation}. \autoref{tab:ablation} ablates the target, gate terms, and encoder. First, the physics-aware target pays off after fine-tuning: adding the texture and saliency channels to the gradient target raises SSDD mAP by 1.9 and WHU-OPT-SAR mIoU by 0.2, but lowers Cls accuracy by 0.85 points. Second, the scatter-aware gain and the physics gate are unreliable individually: added to the physics-aware target, the gate raises Cls accuracy by 1.67 points but lowers SSDD mAP by 3.8, and the gain lowers SSDD mAP by 1.4. Third, the two are complementary: together they raise all three metrics over the physics-aware target alone, by 2.48 points, 0.8~mAP, and 1.8~mIoU, and they also help the gradient target, so the full model is best on every task. Overall, it improves on the gradient-only target of prior SAR models (row 1) by 1.63 points, 2.7~mAP, and 2.0~mIoU. Retaining scattering structures thus takes both the physics-aware target and the joint gate. Replacing the Mamba-3 encoder with a ViT-B with the same pretraining lowers Cls accuracy by 1.4 points, SSDD mAP by 3.1, and WHU-OPT-SAR mIoU by 2.1, so the encoder also contributes to the gains of SARVim.

\begin{table}[t]
  \centering
  \caption{\textbf{Ablation of the reconstruction target, the spatial gate, and the encoder on SARVim-B.} $^*$: ViT-B encoder with the same pretraining.}
  \label{tab:ablation}
  \small
  \setlength{\tabcolsep}{4.5pt}
  \begin{tabular}{@{}clccccc@{}}
    \toprule
        & Target            & Gain $s$  & Gate $\phi$ & Cls                & SSDD              & WHU               \\
    \midrule
    (1) & $G$               &           &             & 92.29              & 71.5              & 45.7              \\
    (2) & $G$               & \ding{51} & \ding{51}   & 93.06              & 72.2              & 45.8              \\
    (3) & $G$\,+\,$T$,\,$S$ &           &             & 91.44              & \underline{73.4}  & 45.9              \\
    (4) & $G$\,+\,$T$,\,$S$ & \ding{51} &             & 91.66              & 72.0              & \underline{46.1}  \\
    (5) & $G$\,+\,$T$,\,$S$ &           & \ding{51}   & \underline{93.11}  & 69.6              & 45.6              \\
    \rowcolor{oursbg}
    (6) & $G$\,+\,$T$,\,$S$ & \ding{51} & \ding{51}   & \textbf{93.92}     & \textbf{74.2}     & \textbf{47.7}     \\
    \midrule
    (7)$^*$ & $G$\,+\,$T$,\,$S$ &           &             & 90.01               & 70.3              & 43.8              \\
    \bottomrule
  \end{tabular}
\end{table}

\begin{figure}[!t]
  \centering
  \includegraphics[width=\columnwidth]{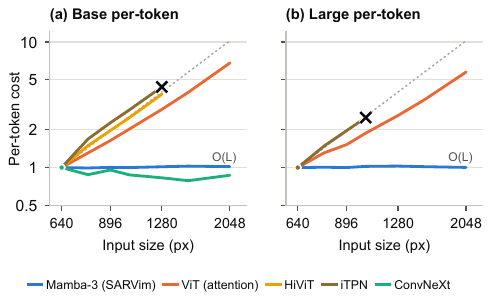}
  \caption{\textbf{Per-token cost versus input size.} Relative to 640-pixel input; crosses mark the largest input that fits in memory.}

  \label{fig:efficiency}
\end{figure}

\subsection{Efficiency}
\label{sec:exp_efficiency}

The per-token cost of SARVim stays within 0.99--1.03 of its 640-pixel value up to 2048 pixels for both sizes (\autoref{fig:efficiency}). That of global-attention encoders grows: on the same stride-8 token grid, width-matched ViTs reach $6.8{\times}$ and $5.8{\times}$ at 2048 pixels, and the HiViT and iTPN-B backbones of SARATR-X and SARATR-X-v2 reach $3.8{\times}$ and $4.4{\times}$ at 1280 pixels before running out of memory. SARVim thus overtakes the matched ViTs between 768 and 896 pixels and is $4.0$--$4.9{\times}$ faster at 2048 pixels. Convolutional and window-attention backbones also scale linearly and run faster (supplementary \autoref{fig:supp_latency}), but mix only local context in each layer and trail SARVim-B in \autoref{tab:sota}. Since SARVim-B also outperforms a ViT-B with the same pretraining on all three tasks, its linear cost comes without a loss of accuracy.

\section{Conclusion}
\label{sec:conclusion}

We present SARVim, a linear-time SAR foundation model pretrained from scratch on information-thin images. Our analysis shows that, in scenes with homogeneous clutter and sparse targets, clutter admits a recursive summary, speckle creates no shared information, and task evidence concentrates on a few scattering structures, as statistics of MillionSAR and four detection benchmarks support. We introduce MillionSAR, 1.07M images from 29 public datasets, enabling pretraining without ImageNet. Following the analysis, SARVim masks tokens by scatter saliency, reconstructs physics-aware targets, emphasizes scatterers with a spatial gate, and uses a Mamba-3 encoder. Experiments show SARVim-B leads Base-scale foundation models on seven of eight benchmarks and SARVim-L competes with Large-scale encoders using little over half their parameters, while the per-token cost stays nearly constant with input size. Emphasizing scattering structures, where information-thin images hold their task information, improves all three tasks. Allocating tokens by scatter saliency could further cut the cost of background. We believe MillionSAR and SARVim will provide a strong basis for SAR foundation models, whose design can consider not only speckle suppression but also where the task information of a SAR image lies.

\section*{LLM Usage Statement}
We used Claude and DeepSeek in the following two capacities.

(1) Writing: we mainly used Claude and DeepSeek to polish the English of the manuscript, and used Claude to cross-check the manuscript against our code for consistency.

(2) Code: we used Claude for routine code writing, error diagnosis, and debugging.

We affirm that we did not use Claude or any other LLM to conceive the research ideas, design the method or experiments, generate or annotate data, or draw conclusions. The authors take full responsibility for the originality, correctness, and integrity of this work.

{
    \small
    \bibliographystyle{ieeenat_fullname}
    \bibliography{main}
}

\clearpage
\makeatletter
\def\NeedsTeXFormat#1[#2]{}
\def\ProvidesPackage#1[#2]{}
\input{stfloats.sty}
\makeatother
\setcounter{page}{1}
\maketitlesupplementary
\appendix

This supplementary material complements the main paper with details and analyses on the proposed SARVim foundation model and the MillionSAR pretraining corpus. The supplementary material is organized as follows:
\begin{itemize}
  \item Section~\ref{sec:supp_proofs}. Proofs of the results on the summary loss, the state size, and the speckle background summary;
  \item Section~\ref{sec:supp_data}. Composition and coverage of the MillionSAR corpus, and its separation from downstream test data;
  \item Section~\ref{sec:supp_impl}. Hyperparameters and implementation details of pretraining and fine-tuning;
  \item Section~\ref{sec:supp_qual}. Extended visualization results for the pretraining targets, segmentation, and detection;
  \item Section~\ref{sec:supp_ssdd}. Additional analyses on SSDD, including a pixel-level study, a grouped detection evaluation, and a speckle sensitivity sweep.
\end{itemize}

\section{Proofs}
\label{sec:supp_proofs}

\paragraph{Summary Loss.} We prove \autoref{prop:summary} and \autoref{eq:state}. For any variable $V$, the Bayes-optimal probabilistic predictor under log-loss is $p(Y\mid V)$, whose expected loss is $H(Y\mid V)$. Since $S$ is a function of $X_A$, $H(Y\mid X_A,X_B)=H(Y\mid X_A,S,X_B)$, and therefore
\begin{equation*}
  \begin{aligned}
    \mathcal{L}^*(S,X_B)-\mathcal{L}^*(X_A,X_B)&=H(Y\mid S,X_B)\\
    &\quad-H(Y\mid X_A,S,X_B)\\
    &=I(Y;X_A\mid S,X_B).
  \end{aligned}
\end{equation*}
For \autoref{eq:state}, in a forward recurrence the outputs after the cut are a function of $(S_t,X_B)$, so the data processing inequality~\cite{cover2006elements} gives $I(X_A;Z_B\mid X_B)\le I(X_A;S_t\mid X_B)$. For a bidirectional or multi-layer model, let $S_t$ be the concatenation of all states that cross the cut (the forward state at the cut in every layer); every suffix output is still a function of $(S_t,X_B)$, and $B_{\mathrm{state}}$ is the total number of bits in $S_t$. Moreover, $I(X_A;S_t\mid X_B)\le H(S_t\mid X_B)\le H(S_t)$, and at finite precision $S_t$ takes at most $2^{B_{\mathrm{state}}}$ values, so $H(S_t)\le B_{\mathrm{state}}$.

\paragraph{Task Information and Pixel Correlation.} Low pixel correlation across the cut does not by itself make a small state sufficient: if $X_A$ and $X_B$ are independent uniform bits and $Y=X_A\oplus X_B$, then $I(X_A;X_B)=0$, yet predicting $Y$ from the suffix still requires one bit from the prefix.

\paragraph{Background Summary, Part (i).} We prove part (i) of \autoref{prop:clutter}. The joint density of $X_1,\dots,X_n$ is
\begin{equation*}
  \prod_{i=1}^{n}\frac{(L/\mu)^L}{\Gamma(L)}X_i^{L-1}e^{-LX_i/\mu},
\end{equation*}
which, for known $n$ and $L$, depends on $\mu$ only through $\sum_iX_i$. By the factorization theorem, $T_n$ is therefore sufficient for $\mu$, and it is updated online by $T_{n+1}=T_n+(1,X_{n+1})$. The sample mean is moreover efficient: the log-density of one sample has second derivative $L/\mu^2-2LX_i/\mu^3$ in $\mu$, so its Fisher information is $L/\mu^2$, and the Cram\'er--Rao bound for unbiased estimators from $n$ samples is $\mu^2/(nL)$. Since $\mathrm{Var}(X_i)=\mu^2/L$, the sample mean $\bar{X}$ is unbiased with $\mathrm{Var}(\bar{X})=\mu^2/(nL)$ and attains the bound. A recurrence with a fixed decay matches only a limited number of samples: in the stationary regime, a fixed exponential average $h_t=\rho h_{t-1}+(1-\rho)X_t$ equals $h_t=(1-\rho)\sum_{k\ge0}\rho^kX_{t-k}$, so
\begin{equation*}
  \begin{aligned}
    \mathrm{Var}(h_t)&=(1-\rho)^2\sum_{k\ge0}\rho^{2k}\,\frac{\mu^2}{L}\\
    &=\frac{\mu^2}{L}\cdot\frac{1-\rho}{1+\rho},
  \end{aligned}
\end{equation*}
which equals $\mu^2/(n_{\mathrm{eff}}L)$, the variance of a sample mean over $n_{\mathrm{eff}}=(1+\rho)/(1-\rho)$ samples.

\paragraph{Background Summary, Part (ii).} We prove part (ii) of \autoref{prop:speckle}. Since $N_A$ and $N_B$ are independent of each other and of $R$, $X_A=R_AN_A$ depends on $(R_B,X_B)$ only through $R_A$, and $X_B=R_BN_B$ depends on $(R_A,X_A)$ only through $R_B$. Hence $X_A\leftarrow R_A\rightarrow R_B\rightarrow X_B$ is a Markov chain, and applying the data processing inequality twice gives $I(X_A;X_B)\le I(R_A;X_B)\le I(R_A;R_B)$. The proof uses only that $X_A$ is a function of $(R_A,N_A)$ and $X_B$ of $(R_B,N_B)$ with $N_A\perp N_B$ and $N_A,N_B\perp R$, so under the same independence assumptions it also covers pixels dominated by deterministic scatterers, which are the special case $N=1$. Part (ii) also implies that more looks never reduce the shared information: for $L'<L$, multiplying each pixel of an $L$-look image by an independent $\tfrac{L}{L'}B_u$ with $B_u\sim\mathrm{Beta}(L',L-L')$ yields an exact $L'$-look image of the same $R$. To see this, write $N_u=G_u/L$ with $G_u\sim\mathrm{Gamma}(L,1)$, and let $B_u\sim\mathrm{Beta}(L',L-L')$ be independent of all other variables. By the beta--gamma algebra, $G_uB_u\sim\mathrm{Gamma}(L',1)$, so $\tfrac{L}{L'}B_uN_u=G_uB_u/L'\sim\mathrm{Gamma}(L',1/L')$, independently across pixels and of $R$. Therefore $X'_u=\tfrac{L}{L'}B_uX_u=R_u\,G_uB_u/L'$ is an exact $L'$-look image of the same $R$. Since $X'_A$ and $X'_B$ are obtained from $X_A$ and $X_B$ through independent channels, $X'_A\leftarrow X_A\rightarrow X_B\rightarrow X'_B$ is a Markov chain, and $I(X'_A;X'_B)\le I(X_A;X_B)$. Thus $I(X_A;X_B)$ is non-decreasing in $L$.

\section{MillionSAR Details}
\label{sec:supp_data}

\begin{table*}[b]
  \centering
  \caption{\textbf{Sources of MillionSAR.} Rows grouped by content. Platform: spaceborne (Space), airborne (Air), or simulation (Sim.). Res.: nominal ground resolution. Size: shorter side, 5th--95th percentile.}
  \label{tab:supp_sources}
  \small
  \renewcommand{\arraystretch}{0.85}
  \setlength{\tabcolsep}{8pt}
  \begin{tabular}{@{}llrcccrr@{}}
\toprule
Source & Content & Year & Platform & Band & Res.\ (m) & Size (px) & Images \\
\midrule
SAR-Ship-Dataset~\cite{wang2019sarship} & Ship & 2019 & Space & C & 3--25 & 256 & 39,729 \\
OpenSARShip~\cite{huang2018opensarship,li2017opensarship2} & Ship & 2017 & Space & C & 2.3--17.4 & 25--197 & 26,679 \\
HRSID~\cite{wei2020hrsid} & Ship & 2020 & Space & C/X & 0.5--3 & 800 & 5,604 \\
FUSAR-Ship 1.0~\cite{hou2020fusar} & Ship & 2020 & Space & C & 1 & 256--512 & 5,243 \\
NASTaR~\cite{hosseiny2026nastar} & Ship & 2025 & Space & S & -- & 512--1024 & 4,801 \\
SSDD~\cite{zhang2021ssdd} & Ship & 2021 & Space & C/X & 1--15 & 253--401 & 1,160 \\
AIR-SARShip~\cite{sun2019airsarship} & Ship & 2019 & Space & C & 1--3 & 512--1000 & 801 \\
SRSDD-v1.0~\cite{lei2021srsdd} & Ship & 2021 & Space & C & 1 & 1024 & 532 \\
\midrule
SAR-AIRcraft-1.0~\cite{wang2023saraircraft} & Aircraft & 2023 & Space & C & 1 & 512 & 18,888 \\
SAR-ACD~\cite{sun2022saracd} & Aircraft & 2022 & Space & C & 1 & 46--112 & 3,032 \\
SADD~\cite{zhang2022sadd} & Aircraft & 2022 & Space & X & 0.5--3 & 224 & 883 \\
SAR-Airport-1.0~\cite{wang2022sarairport} & Aircraft & 2022 & Space & C & 10 & 2048 & 624 \\
\midrule
ATRNet-STAR~\cite{liu2026atrnetstar} & Vehicle & 2025 & Air & X/Ku & 0.12--0.15 & 128 & 198,974 \\
SARSim~\cite{malmgrenhansen2017sarsim} & Vehicle & 2017 & Sim. & X & 0.3 & 139 & 21,168 \\
MSTAR~\cite{ross1998mstar} & Vehicle & 1995 & Air & X & 0.3 & 128--177 & 9,165 \\
SAMPLE~\cite{lewis2019sample} & Vehicle & 2019 & Sim. & X & 0.3 & 128 & 5,380 \\
Sandia MiniSAR~\cite{sandia2006minisar} & Vehicle & 2006 & Air & Ku & 0.1 & 224 & 3,927 \\
SIVED~\cite{lin2023sived} & Vehicle & 2023 & Air & X/Ku/Ka & 0.1--0.3 & 512 & 1,044 \\
\midrule
M4-SAR~\cite{wang2026m4sar} & Multi-class & 2026 & Space & C & 10--60 & 512 & 112,174 \\
RSAR~\cite{zhang2025rsar} & Multi-class & 2025 & Space & C/X & 0.5--10 & 256--512 & 78,837 \\
MSAR~\cite{chen2022msar} & Multi-class & 2022 & Space & C & 1 & 256 & 28,449 \\
OGSOD~\cite{wang2023ogsod} & Multi-class & 2023 & Space & C & 3 & 256 & 18,331 \\
FSAR-Cap~\cite{zhang2026fsarcap} & Multi-class & 2026 & Space & C & -- & 1024 & 14,480 \\
\midrule
SEN1-2~\cite{schmitt2018sen12} & Land cover & 2018 & Space & C & 10 & 256 & 282,384 \\
SARLO-80~\cite{debuysere2026sarlo80} & Land cover & 2025 & Space & X & 0.8 & 1024 & 119,566 \\
OpenSARUrban~\cite{zhao2020opensarurban} & Land cover & 2020 & Space & C & 10 & 100 & 32,868 \\
SARBuD~\cite{wu2022sarbud,wu2021builtup} & Land cover & 2022 & Space & C & 10 & 256 & 25,000 \\
AIR-PolSAR-Seg 2.0~\cite{wang2025airpolsarseg2} & Land cover & 2025 & Space & C & 8 & 512 & 4,932 \\
AIR-PolSAR-Seg 1.0~\cite{wang2022airpolsarseg} & Land cover & 2021 & Space & C & 8 & 512 & 1,400 \\
\midrule
Total & & & & & & & 1,066,055 \\
\bottomrule
\end{tabular}

\end{table*}

\begin{figure*}[t]
  \centering
  \begin{minipage}[t]{0.78\textwidth}
    \centering
    \includegraphics[trim=0 0 95bp 0,clip,width=\linewidth]{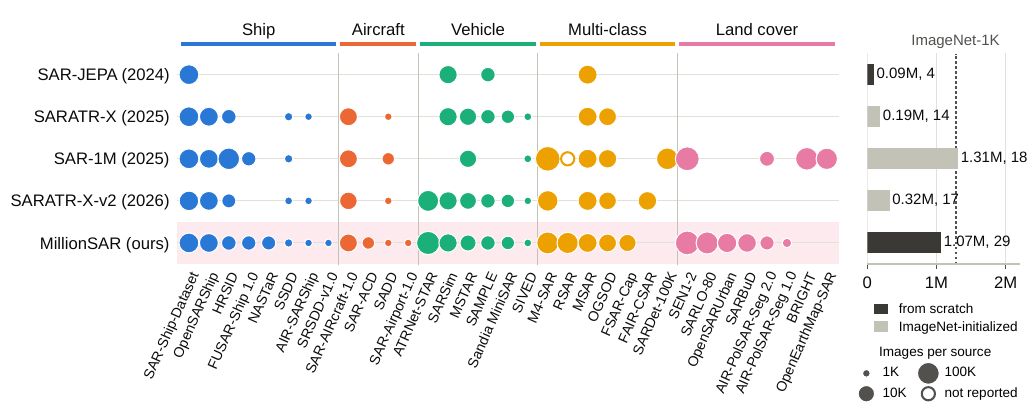}
    \captionof{figure}{\textbf{Composition of SAR pretraining corpora.} Rows: foundation-model corpora; columns: public SAR datasets grouped by content. Dot area grows with log images; white dots: counts not reported.}
    \label{fig:corpora}
  \end{minipage}\hfill
  \begin{minipage}[t]{0.19\textwidth}
    \centering
    \includegraphics[trim=403bp 0 0 0,clip,width=\linewidth]{figs/millionsar/millionsar_corpora.pdf}
    \captionof{figure}{\textbf{Scale of SAR corpora.} }
    \label{fig:corpora_totals}
  \end{minipage}

  \vspace{10pt}

  \begin{minipage}[t]{0.82\textwidth}
    \centering
    \includegraphics[width=\linewidth]{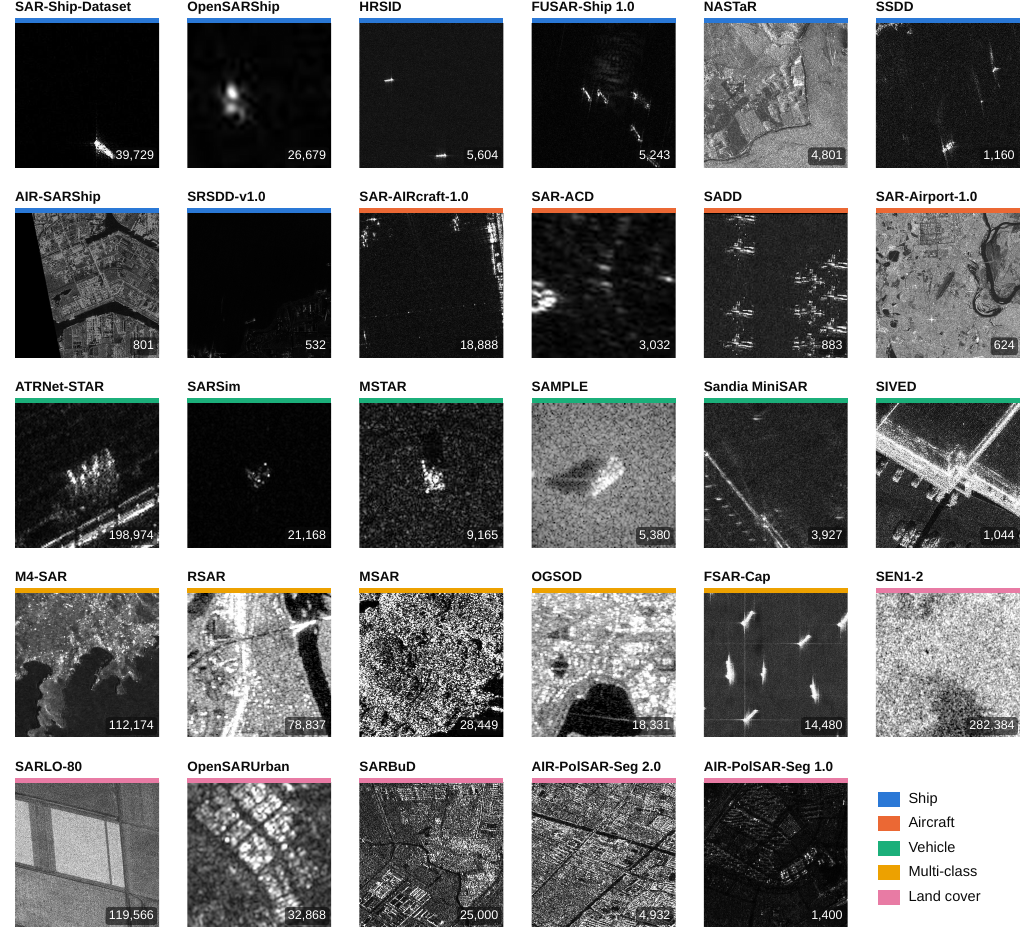}
    \captionof{figure}{\textbf{Example images of the 29 sources.} Grouped by content with image counts. Each example has entropy typical of its source and concentrated structure, so targets remain visible.}
    \label{fig:supp_gallery}
  \end{minipage}
\end{figure*}

MillionSAR aggregates 29 public SAR datasets and 1,066,055 images. By frequency band, 57.7\% of the images are C-band and 14.6\% X-band, with S, Ku, and Ka acquisitions and simulations also represented; resolutions span 0.1--60~m and footprints 17~m--20~km. Only 5.7\% of the images come from datasets released before 2018, and seven sources date from 2025--2026. \autoref{tab:supp_sources} lists each source by content, platform, band, resolution, and image count. \autoref{fig:supp_gallery} shows one example per source, \autoref{fig:corpora} and \autoref{fig:corpora_totals} compare the composition and scale with earlier corpora, and \autoref{fig:pixel} of the main paper summarizes pixel statistics.

\paragraph{Long-Tailed Sources.} The sources follow a long tail: the five largest contribute 74.3\% of the images, SEN1-2 alone 26.5\%, while SRSDD-v1.0 contributes 532 (0.05\%). Fifteen of the 29 sources fall below 1\% (48{,}528 images, 4.6\% jointly), and the largest-to-smallest ratio is 531:1. We sample uniformly across the concatenated sources, retaining this imbalance; pretraining is label-free, so no class prior needs correction.

\paragraph{Comparison with Earlier Corpora.} For comparison, SAR-1M of SARMAE~\cite{liu2025sarmae} contains 1,042,156 SAR--optical pairs from 18 datasets, MuSID of SUMMIT~\cite{du2025summit} 569{,}635 images, and SAR-JEPA~\cite{li2024sarjepa} 94{,}776 images. MillionSAR is comparable in scale but contains only SAR images, with no optical counterpart. Trained from scratch without ImageNet initialization, its representation must come from the SAR clutter statistics and scattering structures of \autoref{sec:theory}, not optical priors. It also aggregates more source datasets, nine of which appear only in its row of \autoref{fig:corpora}.

\paragraph{Test-Set Contamination Check.} For the seven downstream benchmarks of \autoref{sec:exp_setup} overlapping MillionSAR, we remove every image matching an official test split by content hash, so no test image enters pretraining; \autoref{tab:supp_sources} counts source pools before removal. WHU-OPT-SAR is absent entirely, so it cleanly checks leakage: SARVim-B reaches 47.7~mIoU, 1.6 points above SARATR-X-v2 with iTPN-B, reflecting transferable SAR representations rather than overlap with evaluation data.


\clearpage
\makeatletter
\global\@colht\textheight
\global\@colroom\textheight
\makeatother
\raggedbottom

\section{Implementation Details}
\label{sec:supp_impl}

\begin{figure*}[b]
  \centering
  \includegraphics[width=\textwidth]{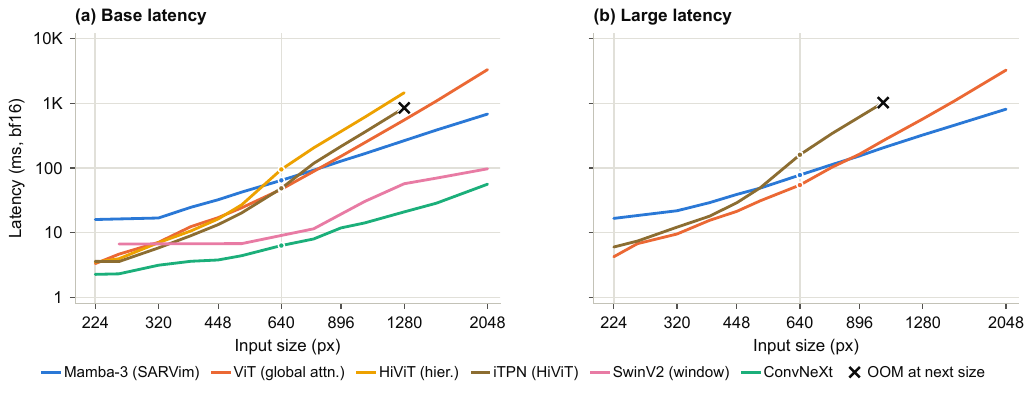}
  \caption{\textbf{Absolute latency versus input size.} Measured on a 16\,GB RTX~5080 (bf16, batch size 1) for the Base \textbf{(a)} and Large \textbf{(b)} models. Crosses mark the largest input that fits in memory. The per-token cost is in \autoref{fig:efficiency}.}
  \label{fig:supp_latency}
\end{figure*}

\paragraph{Experiment Hyperparameters.} We pretrain SARVim-B and SARVim-L on MillionSAR. The encoder has 24 Mamba-3 layers in 12 bidirectional pairs, with widths 768 (B) and 1024 (L) and 96.1M and 168.0M parameters. We train with AdamW~\cite{loshchilov2019adamw} for 600 epochs at a batch size of 4,096 (32 GH200 GPUs $\times$ 128), learning rate $2.4\times10^{-3}$, weight decay 0.05, and bf16, with random resized crops, flips, and rotations but no color augmentation; the schedule uses 30 warmup epochs and cosine decay. For detection, we adopt a unified protocol based on the public code of SARATR-X~\cite{li2025saratrx}, SARMAE~\cite{liu2025sarmae}, and SAMBA~\cite{wang2026samba}, with the detection head and hyperparameters~\cite{lyu2022rtmdet, chen2019mmdetection}. All downstream tasks follow standard evaluation protocols, with configurations released with the code.

\paragraph{Ablation Details.}
The ViT-B row of \autoref{tab:ablation} uses a standard ViT-B, comparable in scale to SARVim-B and trained with the same recipe. For masking, the ablation of \autoref{sec:exp_ablation} borrows SAMBA's comparison between scatter-guided and uniform random masking, which favors scatter-guided. To better validate our method, we additionally pretrain SARVim-B with uniform random masking under the current recipe, keep all other settings of the saliency-guided run, and evaluate on Cls, SSDD, and WHU.

\begin{center}
  \small
  \captionof{table}{\textbf{Random-mask ablation on SARVim-B.} Cls is the linear-probe top-1 accuracy, SSDD is detection mAP, and WHU is WHU-OPT-SAR mIoU.}
  \label{tab:supp_mask}
  \begin{tabular}{@{}lccc@{}}
    \toprule
    Masking & Cls & SSDD & WHU \\
    \midrule
    Uniform random & 92.25 & 69.2 & 44.2 \\
    Saliency-guided (ours) & 93.92 & 74.2 & 47.7 \\
    \bottomrule
  \end{tabular}
\end{center}

\newpage

\autoref{tab:supp_mask} reports the comparison: uniform random masking obtains 92.25 Cls, 69.2 SSDD mAP, and 44.2 WHU-OPT-SAR mIoU, against 93.92, 74.2, and 47.7 for saliency-guided masking, so it trails by 1.7 Cls points, 5.0 AP on SSDD, and 3.5 mIoU on WHU-OPT-SAR. Both variants use the same epochs and images, so the difference isolates the masking rule and shows that saliency-guided masking pretrains the encoder more effectively: target-bearing scattering structures are sparse, so uniform sampling reaches them only by chance, whereas the saliency map selects them directly. Uniform random masking still draws from every image, so the comparison concerns which tokens are reconstructed, not which data are seen.

\paragraph{Width Matching and Absolute Latency.} At equal width and depth, an attention block has about 1.8$\times$ the parameters of a Mamba block, so the matched ViTs in \autoref{fig:efficiency} have widths 576 and 768 at the Base and Large scales, with comparable multiply-accumulates at $224{\times}224$. \autoref{fig:supp_latency} reports the encoder-only latency from 224 to 2048 pixels: at $640{\times}640$ with stride 8, SARVim-B and SARVim-L need 64.5 and 77.9~ms against 46.9 and 54.5~ms for the matched ViTs, because the Mamba-3 scan pays a fixed kernel-launch cost and is less efficient per multiply-accumulate than the ViT's fused attention and matrix-multiplication kernels on this GPU. As the token count grows, the ordering reverses between 768 and 896 pixels; at 2048 pixels, SARVim-B and SARVim-L are 4.0--4.9$\times$ faster than the matched ViTs, and at $2048{\times}2048$, SARVim-B needs 4.9\,GB of peak memory. The fixed scan overhead is thus amortized at large token counts. Convolutional and window-attention backbones remain faster in absolute terms but mix only local context, so the comparison isolates the cost of global mixing.

\newpage

\section{Visualization Results}
\label{sec:supp_qual}

\begin{figure*}[b]
  \centering
  \includegraphics[width=\textwidth]{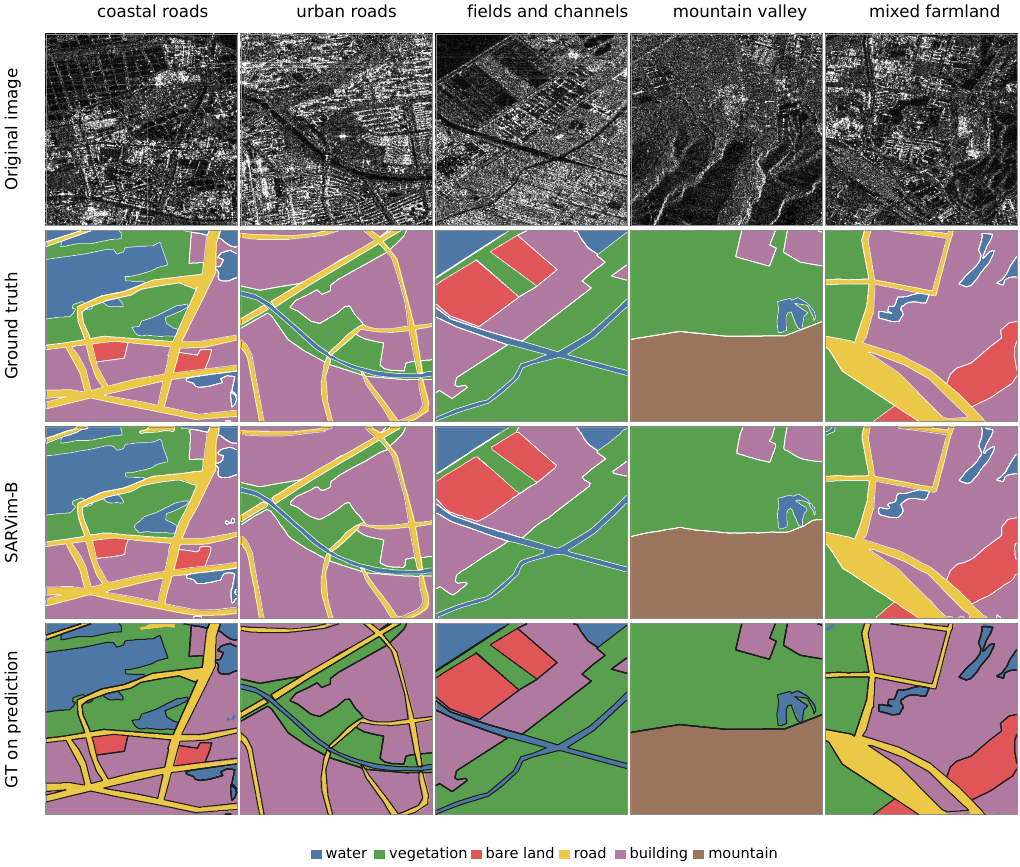}
  \caption{\textbf{Segmentation results of SARVim-B on AIR-PolSAR-Seg-2.0.}}
  \label{fig:supp_qual_seg}
\end{figure*}

\begin{figure*}[!p]
  \centering
  \includegraphics[width=0.83\textwidth]{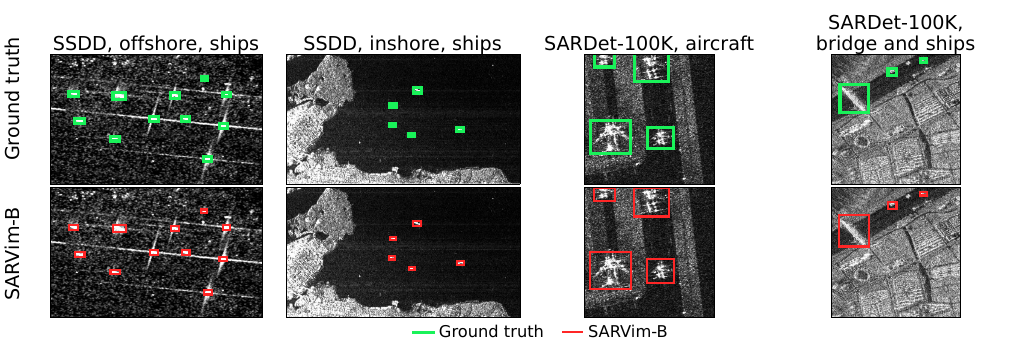}
  \caption{\textbf{Detection results of SARVim-B on SSDD and SARDet-100K.}}
  \label{fig:supp_qual_det}
\end{figure*}

\autoref{fig:supp_qual_seg} and \autoref{fig:supp_qual_det} visualize the results of the fine-tuned SARVim-B on segmentation and detection; Tables~\ref{tab:sota} and~\ref{tab:clsseg} report the aggregate performance. In \autoref{fig:supp_qual_seg}, five AIR-PolSAR-Seg-2.0 validation patches are columns and the rows give the original image, ground truth, prediction, and ground-truth boundaries in dark, so a missed or displaced boundary shows as a dark line in a uniform region. In \autoref{fig:supp_qual_det}, the upper row is ground truth and the lower row predictions at score threshold 0.3: columns 1--2 are SSDD images with 11 offshore and 5 inshore ships, and columns 3--4 SARDet-100K images with four aircraft and one bridge with two ships, all detected at IoU 0.5 without false positives.

\begin{figure*}[!p]
  \centering
  \includegraphics[width=0.72\textwidth]{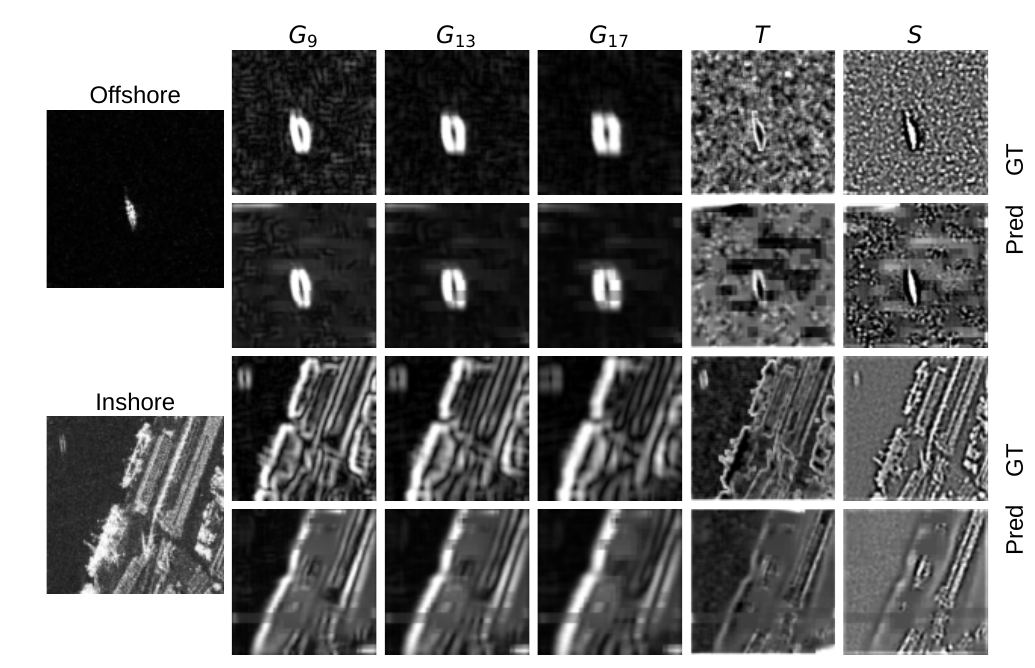}
  \caption{\textbf{Physics-aware targets and reconstructions of SARVim-B.}}
  \label{fig:supp_recon}
\end{figure*}

\begin{figure*}[!p]
  \centering
  \includegraphics[width=0.72\textwidth]{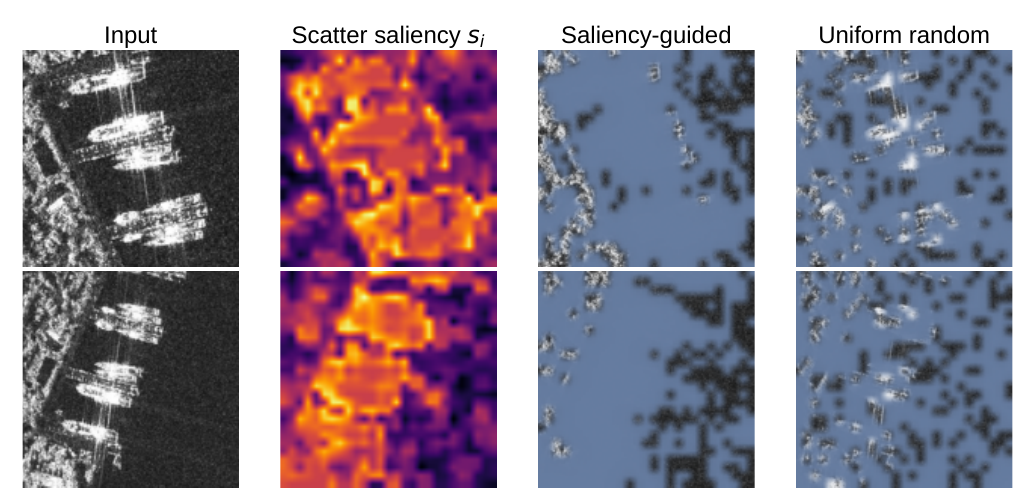}
  \caption{\textbf{Saliency-guided versus random masking.}}
  \label{fig:supp_mask}
\end{figure*}

\paragraph{Pretraining Targets.} \autoref{fig:supp_recon} shows, for an offshore and an inshore image, the five physics-aware targets and the decoder predictions at 75\% masking: the ship layout and the harbor outline survive in $G_9$, $G_{13}$, $G_{17}$, and $S$, while $T$ is predicted as a smoother field than its speckled target, consistent with the local-statistics target of \autoref{sec:method_pretrain}. \autoref{fig:supp_mask} contrasts the saliency-guided mask with uniform random masking at the same 75\% ratio on two inshore images: both masks keep the same 25\% of tokens visible, so the difference is only in placement---the guided mask concentrates the removals on the ship, its wake, and the harbor structures, whereas random masking spreads them uniformly and leaves much of each target visible.

\makeatletter
\setlength{\@dblfptop}{0pt plus 1fil}
\setlength{\@dblfpsep}{8pt}
\setlength{\@dblfpbot}{0pt plus 1fil}
\makeatother

\clearpage

\clearpage

\section{SSDD: Pixel Study and Grouped Detection}
\label{sec:supp_ssdd}

This section complements \autoref{sec:exp_theory}'s detection study. We ask which pixel statistics carry ship evidence on SSDD, then group detections by scene, density, and box size.

\paragraph{Images and Metrics.}
We use two official SSDD test lists~\cite{zhang2021ssdd}: 186 offshore and 46 inshore images, each an 8-bit JPEG; gray values are relative intensities, not calibrated backscatter, and the box union marks background inside rectangles as positive, so AUCs are lower bounds. We rank pixels by gray value, the $3{\times}3$ mean, and its log ratios to a four-direction scalar recurrence background and a $9{\times}9$ ring mean; recall uses the top 10\% by the $3{\times}3$ mean, and intervals are image-level bootstraps.

\begin{figure*}[t]
  \centering
  \includegraphics[width=\textwidth]{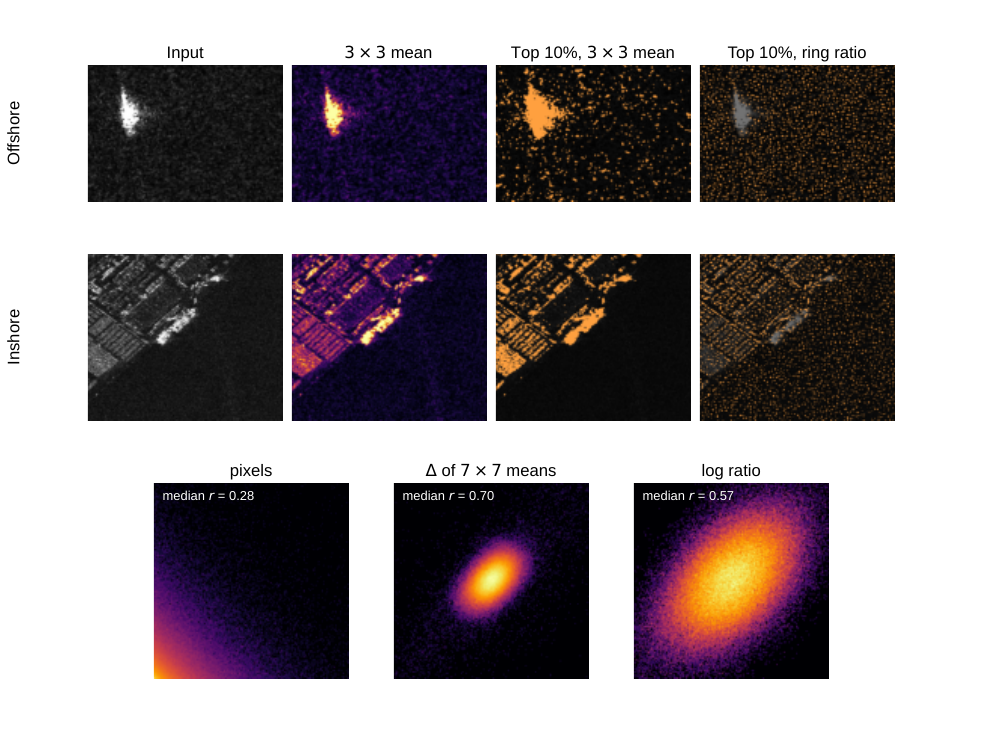}
  \caption{\textbf{Pixel-level selection and speckle correlation on SSDD.} Top: 10\% highest pixels by $3{\times}3$ mean and $9{\times}9$ ring ratio (amber). Bottom: joint density of two Gamma(1) copies of a representative image; panel values give the median Pearson $r$ over 232 images.}
  \label{fig:supp_pixel}
\end{figure*}

\paragraph{Pixel-Level Findings.} The top rows of \autoref{fig:supp_pixel} show, for an offshore and an inshore image, the input, the $3{\times}3$ mean, and the top 10\% pixels by this mean and by the $9{\times}9$ ring ratio, in amber. Offshore the mean's selection covers the ship, so the evidence follows property (iii); inshore bright harbor structures take much of it and only about a third of ship pixels remain, with recall 77.6\% offshore and 35.5\% inshore (\autoref{tab:ssdd}), and the mean separates box pixels with 95\% intervals [0.900, 0.912] offshore and [0.710, 0.787] inshore, both above 0.5. Ring-ratio selections scatter over background in both rows, with AUC 0.490 offshore and 0.501 inshore. A fixed averaged recurrence is not enough: it lowers the AUC by a median of 0.281 and 0.153 relative to the $3{\times}3$ mean, while ring windows from $5{\times}5$ to $21{\times}21$ stay between 0.47 and 0.53, so what to keep is better learned from data (\autoref{sec:exp_theory}). The bottom row of \autoref{fig:supp_pixel} multiplies one image by two independent Gamma(1) speckle fields and plots the joint density of the noisy copies for three statistics; a brighter cloud hugging the diagonal means the copies agree. Raw pixels give a diffuse cloud, with median Pearson $r$ 0.280; the bright corner is the pile-up of near-zero values, not correlation. The difference of adjacent $7{\times}7$ means and their log ratio stay compact, at 0.702 and 0.567, because neighborhood averaging and ratios suppress the multiplicative factor; local statistics are more stable under speckle, supporting the continuous scoring, learned gating, and statistics-based targets of \autoref{sec:method}.

\paragraph{Grouped Detection Evaluation.}\label{sec:supp_groups}
Following \autoref{sec:exp_theory}, we keep every test image and box; blocks (a)--(c) pool scenes or densities, cross them at matched counts, and bin by box size. The bootstrap resamples images 100 times for 95\% intervals; the full set gives 74.2 AP (\autoref{tab:supp_ssdd_groups}). In the crowded inshore cell, AP$_{50}$ stays at 95.2 while AP drops to 65.7: the loss is in ranking, not localization, and the cue is local while the failures are crowded, so we mask by neighborhood statistics and evaluate per density.

\begin{figure*}[t]
  \centering
  \begin{minipage}[t]{0.49\textwidth}
    \centering
    \includegraphics[width=\linewidth]{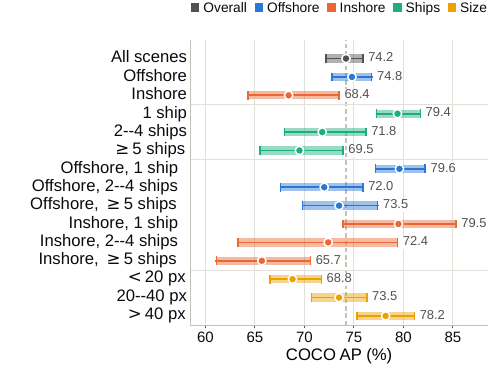}\\[-4pt]
    {\small\textbf{(a)}}
  \end{minipage}\hfill
  \begin{minipage}[t]{0.49\textwidth}
    \centering
    \includegraphics[width=\linewidth]{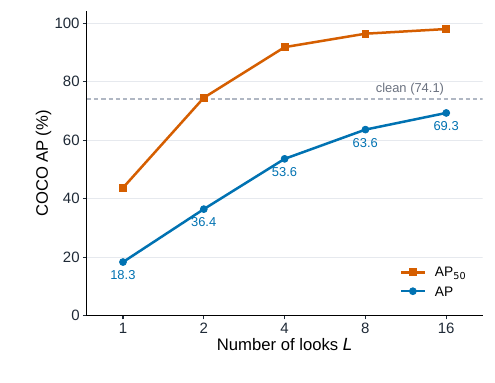}\\[-4pt]
    {\small\textbf{(b)}}
  \end{minipage}
  \caption{\textbf{Grouped AP and speckle sensitivity on SSDD.} (a) COCO AP with 95\% bootstrap intervals for the groups of \autoref{tab:supp_ssdd_groups}; dashed: overall AP. (b) COCO AP and AP$_{50}$ under multiplicative speckle of $L$ looks; dashed: clean AP.}
  \label{fig:supp_groups}
  \label{fig:supp_speckle}
\end{figure*}

\paragraph{Grouped AP.} \autoref{fig:supp_groups}(a) plots the intervals of \autoref{tab:supp_ssdd_groups}. Single-ship AP is 79.6 offshore and 79.5 inshore, effectively the same, but the scenes separate as ships crowd: with five or more ships, inshore AP is 65.7, 7.8 below offshore; box size adds a second gradient, 68.8 below 20~px against 78.2 above 40~px. Adjacent intervals mostly overlap, so we read the ordering of groups, not individual gaps.

\paragraph{Speckle Sensitivity.} To probe sensitivity to speckle statistics, we multiply every test image by an $L$-look Gamma field with unit mean and variance $1/L$, and rerun the trained detector without retraining; the images already contain real speckle, so this adds a second, synthetic layer. \autoref{fig:supp_speckle}(b) shows AP recovering monotonically with $L$, from 18.3 at $L{=}1$ through 36.4, 53.6, and 63.6 at $L{=}2$, 4, and 8, to 69.3 at $L{=}16$, against 74.1 clean, while AP$_{50}$ is already 91.8 at $L{=}4$ and 98.0 at $L{=}16$. At $L{=}1$ the Gamma field has deep nulls that erase much of the target return, an input far outside the training distribution, so the collapse is expected.


\begin{center}
  \begin{minipage}{\columnwidth}
  \centering
  \small
  \captionof{table}{\textbf{Pixel-level statistics on SSDD (medians over images).}}
  \label{tab:ssdd}
  \begin{tabular}{@{}lcc@{}}
    \toprule
                                         & Offshore       & Inshore        \\
    \midrule
    Box-union area                       & 1.4\%          & 2.3\%          \\
    Gray mass in top 10\% of pixels      & 23.1\%         & 30.7\%         \\
    \midrule
    AUC, brightness                      & 0.836          & 0.696          \\
    AUC, $3{\times}3$ mean               & \textbf{0.906} & \textbf{0.750} \\
    Recall in top 10\%, $3{\times}3$ mean & 77.6\%         & 35.5\%         \\
    AUC, fixed recurrent ratio           & 0.613          & 0.572          \\
    AUC, fixed ring ratio                & 0.490          & 0.501          \\
    \bottomrule
  \end{tabular}
  \end{minipage}
\end{center}

\vspace{0pt}

\begin{center}
  \begin{minipage}{\columnwidth}
  \centering
  \small
  \captionof{table}{\textbf{Grouped detection results of SARVim-B on SSDD.}}
  \label{tab:supp_ssdd_groups}
  \footnotesize
  \renewcommand{\arraystretch}{0.92}
  \setlength{\tabcolsep}{2pt}
  \resizebox{\columnwidth}{!}{%
  \begin{tabular}{@{}llrrcc@{}}
    \toprule
    Scene & Group & Images & Boxes & AP [95\% CI] & AP$_{50}$ \\
    \midrule
    \multicolumn{6}{@{}l}{\textit{(a) Marginal scene and density}} \\
    All & -- & 232 & 546 & 74.2 [72.2, 75.9] & 98.5 \\
    Offshore & all & 186 & 374 & 74.8 [72.8, 76.8] & 98.8 \\
    Inshore & all & 46 & 172 & 68.4 [64.3, 73.5] & 95.9 \\
    -- & 1 ship & 135 & 135 & 79.4 [77.3, 81.7] & 100.0 \\
    -- & 2--4 ships & 69 & 174 & 71.8 [68.0, 76.2] & 96.8 \\
    -- & $\ge$5 ships & 28 & 237 & 69.5 [65.5, 73.9] & 97.4 \\
    \midrule
    \multicolumn{6}{@{}l}{\textit{(b) Scene $\times$ density}} \\
    Offshore & 1 ship & 120 & 120 & 79.6 [77.2, 82.2] & 100.0 \\
    Offshore & 2--4 ships & 51 & 128 & 72.0 [67.6, 75.9] & 97.5 \\
    Offshore & $\ge$5 ships & 15 & 126 & 73.5 [69.8, 77.4] & 100.0 \\
    Inshore & 1 ship & 15 & 15 & 79.5 [73.9, 85.3] & 100.0 \\
    Inshore & 2--4 ships & 18 & 46 & 72.4 [63.3, 79.4] & 96.5 \\
    Inshore & $\ge$5 ships & 13 & 111 & 65.7 [61.1, 70.6] & 95.2 \\
    \midrule
    \multicolumn{6}{@{}l}{\textit{(c) Target size (bounding-box side)}} \\
    All & $<$20 px & 60 & 211 & 68.8 [66.5, 71.7] & -- \\
    All & 20--40 px & 117 & 192 & 73.5 [70.7, 76.3] & -- \\
    All & $>$40 px & 105 & 143 & 78.2 [75.3, 81.1] & -- \\
    \bottomrule
  \end{tabular}}
  \end{minipage}
\end{center}

\clearpage

\makeatletter
\setlength{\@dblfptop}{0pt}
\makeatother

\end{document}